\documentclass{article}
\usepackage[utf8]{inputenc}
\usepackage[T1]{fontenc}    
\usepackage{amsmath, amssymb, amsthm, mathtools}
\usepackage{dsfont} 
\usepackage[dvipsnames]{xcolor}
\usepackage{enumitem}
\usepackage{wrapfig}
\usepackage{multirow}
\usepackage{fancybox}
\usepackage{changepage}
\usepackage{siunitx}
\usepackage{natbib}
\usepackage{subcaption}
\usepackage{booktabs}

\usepackage{hyperref} 
\hypersetup{
    colorlinks,
    linkcolor={blue!60!black},
    citecolor={cyan!80!black},
    urlcolor={magenta!60!black}
}
\usepackage[noabbrev, capitalise, nameinlink,]{cleveref}

\theoremstyle{plain}
\newtheorem{theorem}{Theorem}[section]

\theoremstyle{definition}
\newtheorem{definition}[theorem]{Definition}

\newtheorem{remark}[theorem]{Remark}

\usepackage{pgfplots}
\usepackage{pgfplotstable}
\usepgfplotslibrary{groupplots} 
\usepgfplotslibrary{fillbetween}
\usetikzlibrary{positioning}
\pgfplotsset{compat=1.16}
\pgfplotsset{grid = major, grid style={gray!30!white}}
\definecolor{darkgreen}{rgb}{0.31, 0.47, 0.26}
\tikzset{relStyle/.style={blue, mark=*,mark options={fill=blue}}}
\tikzset{absStyle/.style={red, mark=square*,mark options={fill=red}}}
\tikzset{epsiStyle/.style={darkgreen, mark=triangle*,mark options={fill=darkgreen}}}

\tikzset{relStyleKnap/.style={MidnightBlue, mark=*,mark options={fill=MidnightBlue}}}
\tikzset{absStyleKnap/.style={BrickRed, mark=square*,mark options={fill=BrickRed}}}
\tikzset{epsiStyleKnap/.style={OliveGreen, mark=triangle*,mark options={fill=OliveGreen}}}

\tikzset{true_regretStyle/.style={purple, mark=*,mark options={fill=purple}}}
\tikzset{predicted_regretStyle/.style={orange, mark=square*,mark options={fill=orange}}}

\tikzset{reconstStyle/.style={teal, mark=triangle*,mark options={fill=teal}}}

\tikzset{epsproxStyle/.style={brown, mark=triangle*,mark options={fill=brown}}}

\tikzset{losscavaeStyle/.style={violet, mark=*,mark options={fill=violet}}}

\tikzset{losssphereStyle/.style={cyan, mark=*,mark options={fill=cyan}}}

\tikzset{lossloglkStyle/.style={olive, mark=square*,mark options={fill=olive}}}

\DeclareMathOperator{\Exp}{\mathbb{E}}
\newcommand{\xAlt}{{x^{\text{alt}}}}
\newcommand{\yAlt}{{y^{\text{alt}}}}
\newcommand{\thetaAlt}{{\theta^{\text{alt}}}}
\newcommand{\zAlt}{{z^{\text{alt}}}}

\def\argmin{\mathop{\rm argmin}}

\def\min{\mathop{\rm min}}
\def\max{\mathop{\rm max}}
\def\subto{{\rm s.\rm t.}}
\DeclareMathOperator{\DKL}{D_{\text{KL}}}
\def\reg{\mathop{\rm reg}}

\newlist{inlinelist}{enumerate*}{1}
\setlist[inlinelist]{label=(\roman*)} 
\newcommand{\model}[1]{\texttt{#1}}

\usepackage{etoolbox}
\makeatletter
\pretocmd{\NAT@citex}{%
  \let\NAT@hyper@\NAT@hyper@citex
  \def\NAT@postnote{#2}%
  \setcounter{NAT@total@cites}{0}%
  \setcounter{NAT@count@cites}{0}%
    \forcsvlist{\stepcounter{NAT@total@cites}\@gobble}{#3}}{}{}
\newcounter{NAT@total@cites}
\newcounter{NAT@count@cites}
\def\NAT@postnote{}
\def\NAT@hyper@citex#1{%
  \stepcounter{NAT@count@cites}%
  \hyper@natlinkstart{\@citeb\@extra@b@citeb}#1%
  \ifnumequal{\value{NAT@count@cites}}{\value{NAT@total@cites}}
    {\ifNAT@swa\else\if*\NAT@postnote*\else%
     \NAT@cmt\NAT@postnote\global\def\NAT@postnote{}\fi\fi}{}%
  \ifNAT@swa\else\if\relax\NAT@date\relax
  \else\NAT@@close\global\let\NAT@nm\@empty\fi\fi
  \hyper@natlinkend}
\renewcommand\hyper@natlinkbreak[2]{#1}
\patchcmd{\NAT@citex}
  {\ifNAT@swa\else\if*#2*\else\NAT@cmt#2\fi
   \if\relax\NAT@date\relax\else\NAT@@close\fi\fi}{}{}{}
\patchcmd{\NAT@citex}
  {\if\relax\NAT@date\relax\NAT@def@citea\else\NAT@def@citea@close\fi}
  {\if\relax\NAT@date\relax\NAT@def@citea\else\NAT@def@citea@space\fi}{}{}
\makeatother

\usepackage[accepted]{icml2024}

\newcommand{\mytitle}{CF-OPT: Counterfactual Explanations for Structured Prediction}

\icmltitlerunning{\mytitle}

\begin{document}
\twocolumn[
    \icmltitle{\mytitle}

    \begin{icmlauthorlist}
        \icmlauthor{Germain Vivier-Ardisson}{polymagi,cermics}
        \icmlauthor{Alexandre Forel}{polymagi}
        \icmlauthor{Axel Parmentier}{cermics}
        \icmlauthor{Thibaut Vidal}{polymagi}
    \end{icmlauthorlist}
    
    \icmlaffiliation{polymagi}{CIRRELT \& SCALE-AI Chair in Data-Driven Supply Chains, Department of Mathematical and Industrial Engineering, Polytechnique Montreal, Montreal, Canada}
    
    \icmlaffiliation{cermics}{CERMICS, \'Ecole des Ponts, Marne-la-Vall\'ee, France}

    \icmlcorrespondingauthor{Germain Vivier-Ardisson}{germain.vivier-ardisson@enpc.fr}

    \icmlkeywords{explainable AI, end-to-end optimization}
    \vskip 0.3in
]

\printAffiliationsAndNotice{}

\begin{abstract}
    Optimization layers in deep neural networks have enjoyed a growing popularity in structured learning, improving the state of the art on a variety of applications. 
    Yet, these pipelines lack interpretability since they are made of two opaque layers: a highly non-linear prediction model, such as a deep neural network, and an optimization layer, which is typically a complex black-box solver.
    Our goal is to improve the transparency of such methods by providing counterfactual explanations.
    We build upon variational autoencoders a principled way of obtaining counterfactuals: working in the latent space leads to a natural notion of plausibility of explanations.
    We finally introduce a variant of the classic loss for VAE training that improves their performance in our specific structured context.
    These provide the foundations of \texttt{CF-OPT}, a first-order optimization algorithm that can find counterfactual explanations for a broad class of structured learning architectures.
    Our numerical results show that both close and plausible explanations can be obtained for problems from the recent literature.
\end{abstract}
\section{Introduction}
\label{sec:introduction}
Recent studies have shown a surge of interest in developing structured learning pipelines that combine machine learning and optimization \citep{Amos2017optnet, Donti2017}. From the perspective of structured learning, optimization layers translate a machine-learning prediction into an output that maximizes a given objective while respecting structural constraints. From an optimization perspective, machine-learning models can identify patterns in data to optimize an objective function subject to uncertain parameters. While these pipelines have shown great results in many applications \citep{Mandi2023decision, Sadana2023survey}, it is hard to justify their outputs since both their prediction and optimization components are complex and opaque. This lack of interpretability is hardly tackled, even though it is likely to hinder the adoption of this type of solution. We aim to explain the decision-making process of structured learning pipelines by providing counterfactual explanations. Counterfactual explanations show the minimum changes in a feature vector needed to lead the pipeline to output a different decision. By contrasting the initial factual instance with its counterfactual explanation, they allow users to understand the impact of features on the decision. Indeed, identifying which features make a specific solution optimal and establishing a link between changes in the features and changes in the output are essential conditions for the reliable use of a model.
\begin{figure}[ht!]
    \centering
    \vspace{-1mm}
    \begin{subfigure}{0.48\columnwidth}
        \includegraphics[width=\linewidth]{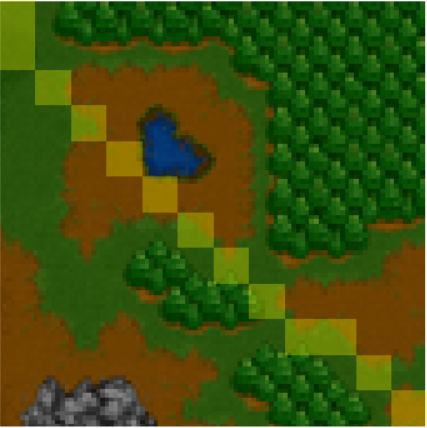}
        \caption{Initial map}
    \end{subfigure}
    \hfill
    \begin{subfigure}{0.48\columnwidth}
        \includegraphics[width=\linewidth]{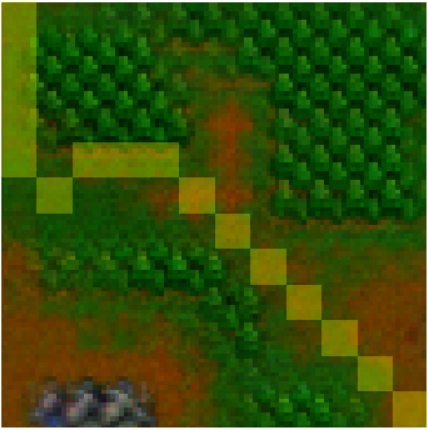}
        \caption{Counterfactual map}
    \end{subfigure}
    \vspace{-2mm}
    \caption{(a)~Initial and (b)~counterfactual maps with their respective shortest path (initial and alternative solutions) shown in yellow. The explanation is given by \model{CF-OPT}. The corresponding pipeline and experiment is detailed in \cref{sec:exp_warcraft} and will serve as a guiding example in this work.}
    \label{fig:mapExample}
    \vspace{-4mm}
\end{figure}

In this paper, we present \model{CF-OPT}, a method to obtain counterfactual explanations for any deep learning model concluded by a black-box linear optimization layer, called thereafter structured learning pipeline. 
It is based on differentiating a relaxed version of the explanation problem and applying a first-order optimization method to obtain a counterfactual. An example explanation provided by \model{CF-OPT} is shown in \cref{fig:mapExample} for the shortest paths on Warcraft maps structured learning pipeline, taken from \citet{vlastelica2019diff}. \newline

We make the following contributions: 
\begin{enumerate}[topsep=-0.8\parskip, itemsep=-3pt, wide=5pt, label=(\roman*)]
    \item We present an efficient algorithm to provide counterfactual explanations to structured learning pipelines that combine a deep model and an optimization model. It is based on an augmented Lagrangian relaxation of the explanation problem and a first-order optimization algorithm.
    \item We propose a new type of explanation that does not require a target decision and can be used for local sensitivity analyses.
    \item We highlight the sensitivity of structured learning pipelines to subtle perturbations: for high-dimensional inputs, naive counterfactual explanations are equivalent to adversarial attacks. As a consequence, we define a notion of plausibility region and develop a method to obtain plausible explanations even in high-dimensional settings thanks to a Variational Auto-Encoder (VAE).
    \item We take advantage of the prior imposed on the latent space of the VAE in order to add a plausibility regularization term to the explanation objective. Furthermore, we modify the training objective of the VAE by introducing a cost-aware loss, designed to account for the downstream optimization task.
    \item We demonstrate the value of our method and analyze its principal components thanks to numerical experiments. We show that we can compute explanations efficiently, even for large problems.
\end{enumerate}

\section{Problem Statement}
\label{sec:problem}
\begin{figure}[htb]
    \centering
    \resizebox{0.98\linewidth}{!}{\begin{tikzpicture}
 
 \node (A) at (0, 0) {};
 
\node [ draw,
        minimum width=2cm,
        minimum height=1.2cm,
        right=2cm of A,
        align=center
       ]  (ml) {Prediction\\model $\varphi$};
 
\node [draw,
        minimum width=2cm, 
        minimum height=1.2cm,
        right=2.5cm of ml,
        align=center
      ] (opt) {Optimization \\ model};

\draw[-stealth] (A) -- node[label= above:Covariate] {} node[label= below:$x$] {} (ml.west);

\draw[-stealth] (ml.east) -- node[label= above:Intermediate] {} node[label= below:{input $\theta$}] {} (opt.west);

\draw[-stealth] (opt.east) -- node[label= above:Decision] {} node[label= below:$y^*$] {} ++ (2,0);

\end{tikzpicture}}
    \vspace{-2mm}
    \caption{Structured learning pipeline.}
    \label{fig:learn_opt_pipeline}
\end{figure}
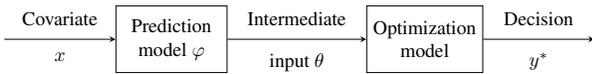

In this paper, we consider structured learning models of the form
\begin{equation}
    \label{eq:model}
    x \longmapsto y^*(x) \in \argmin_{y\in\mathcal{Y}}\varphi(x)^\top y,
\end{equation}
where \text{$x \in \mathcal{X}\subseteq \mathbb{R}^{n_x}$} is a vector of features, equivalently referred to as \textit{context} or \textit{covariate}. The prediction model \text{$\varphi:\mathcal{X} \rightarrow \Theta$} transforms $x$ into an intermediate input \text{$\varphi(x)=\theta \in \Theta \subseteq \mathbb{R}^{n_\theta}$}. This intermediate input parameterizes the linear optimization layer, which returns the decision \text{$y^* \in \argmin_{y \in \mathcal{Y}} \theta^\top y,$} where $\mathcal{Y}\subseteq \mathbb{R}^{n_y}$ is the feasible set. A schematic representation is given in \cref{fig:learn_opt_pipeline}.

We focus on optimization layers of feasible set $\mathcal{Y}$ defined by linear with possible integrality constraints, and make no assumption on the solver used to compute $y^*$.

This setting encompasses a wide array of methods proposed in recent years, used to make structured predictions \citep{Wilder2019, vlastelica2019diff, Ferber2020, Berthet2020, Niepert2021implicit, Sahoo2023backpropagation, Stewart2023differentiable}, to solve contextual stochastic problems whose objective is linear in the uncertain parameters \citep{Elmachtoub2022, Jeong2022exact, McKenzie2023faster}, or to approximate hard optimization problems by learning linear surrogate models \citep{Dalle2022, Ferber2023surco, gupta2024data}. While these works differ in the way the models are trained, the structure of the resulting pipeline, in particular the optimization layer, is identical to the one in \cref{fig:learn_opt_pipeline}. This work thus allows to improve the interpretability of a large array of recent methods. 
The only assumption we make on the prediction model is that it is differentiable w.r.t to its input, so that $\varphi$ can typically be any state-of-the-art deep learning model.
Hence, our setting extends the one of \citet{Forel2023}, who provide explanations for a restricted type of pipeline based on random forests and $k$-nearest neighbors.

\begin{figure*}[ht!]
    \centering
    \begin{minipage}{0.22\textwidth}
        \begin{subfigure}{\linewidth}
            \includegraphics[width=\linewidth]{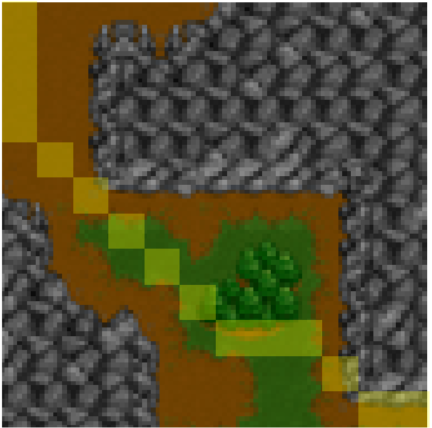}
            \caption{Initial map and its associated shortest path}
            \label{fig:adv1}
        \end{subfigure}
    \end{minipage}\hfill
    \begin{minipage}{0.22\textwidth}
        \begin{subfigure}{\linewidth}
            \includegraphics[width=\linewidth]{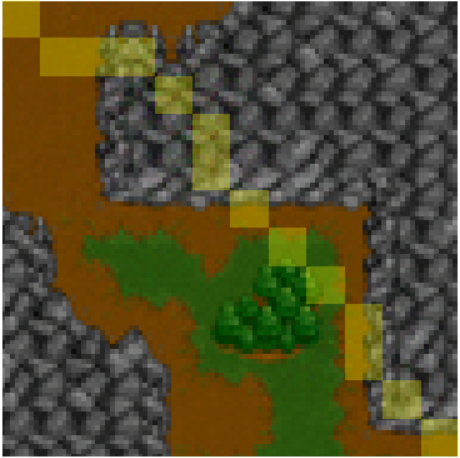}
            \caption{Adversarial explanation obtained with naive algorithm}
            \label{fig:adv2}
        \end{subfigure}
    \end{minipage}\hfill
    \begin{minipage}{0.22\textwidth}
        \begin{subfigure}{\linewidth}
            \includegraphics[width=\linewidth]{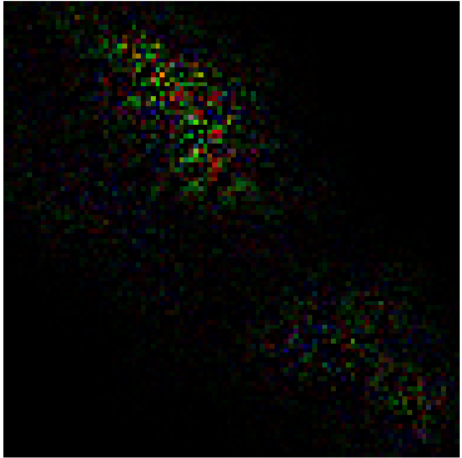}
            \caption{Magnified difference (x10) between (a) and (b)}
            \label{fig:adv3}
        \end{subfigure}
    \end{minipage}\hfill
    \begin{minipage}{0.22\textwidth}
        \begin{subfigure}{\linewidth}
            \includegraphics[width=\linewidth]{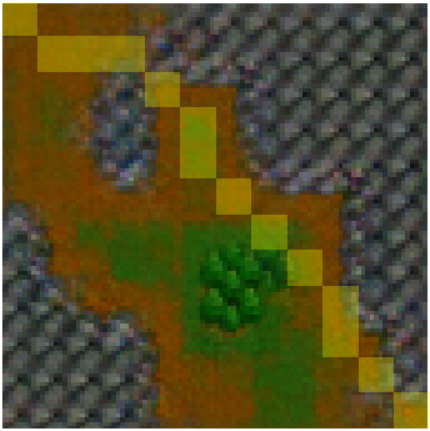}
            \caption{Plausible explanation obtained with \texttt{CF-OPT}}
            \label{fig:adv4}
        \end{subfigure}
    \end{minipage}
    \vspace{-2mm}
    \caption{Naive counterfactual search in raw feature space leads to adversarial examples. \texttt{CF-OPT} recovers a plausible explanation, using a VAE trained in a cost-aware fashion, and a latent hypersphere plausibility regularized objective ($\alpha=2, \beta=10$).}
    \vspace{-3mm}
    \label{fig:adv}
\end{figure*}

\subsection{Counterfactual Explanations of Decisions}
\label{sec:cf_decisions}
Counterfactual explanations of decisions are based on contrasting a covariate with an alternative one. Suppose that, at decision time, we observe covariate~$x_0$ and evaluate the trained pipeline to obtain decision~$y_0$. Let $\yAlt$ be an alternative decision representing, for instance, the solution usually chosen by an expert in context~$x_0$. We explain the decision~$y_0$ by comparing its covariate $x_0$ to an alternative covariate $\xAlt$ that we find, i.e., the counterfactual explanation. Let $\thetaAlt = \varphi(\xAlt)$ be the output of the prediction model for this alternative covariate. Two types of explanation have been introduced by \citet{Forel2023}.

\begin{definition}
    \label{def:rel_exp}
    A covariate $\xAlt$ is a \textbf{\textit{relative explanation}} if the decision $\yAlt$ is at least as good as $y_0$ in the context $\xAlt$, that is: \; $ \smash{\thetaAlt^\top \yAlt \le \thetaAlt^\top y_0}$.
\end{definition}

\begin{definition}
    \label{def:abs_exp}
    A covariate $\xAlt$ is an \textbf{\textit{absolute explanation}} if the decision $\yAlt$ is optimal in the context $\xAlt$, that is: \; $\smash{\yAlt \in \argmin_{y\in\mathcal{Y}}\thetaAlt^\top y}$.
\end{definition}

Definitions~\ref{def:rel_exp} and \ref{def:abs_exp} require that the practitioner provides an alternative solution $\yAlt$. When such a solution is not available, the decision-maker may be interested in analyzing the local sensitivity of the decision $y_0$. To this end, we propose a third, novel type of explanation, aiming at finding a context $\xAlt$ in which $y_0$ is not optimal anymore. To control the amount of suboptimality of the initial solution $y_0$ in the new context $\xAlt$, we add a parameter $\varepsilon$.
\begin{definition}
    A covariate $\xAlt$ is an \textbf{\textit{$\varepsilon$-explanation}} if $\smash{\min_{y\in\mathcal{Y}}\thetaAlt^\top y + \varepsilon\,|\min_{y\in\mathcal{Y}}\thetaAlt^\top y| \le \thetaAlt^\top y_0}$, that is, $y_0$ has a relative optimality gap (or relative regret) of at least $\varepsilon$ in the context $\xAlt$.
\end{definition}

\begin{remark}
\label{positive_cost}
    For the sake of clarity, we will consider below that the costs of solutions are necessarily positive (which is true, e.g., in our guiding example on Warcraft maps). The condition for $\xAlt$ being an $\varepsilon$-explanation then reduces to $\smash{(1+\varepsilon)\min_{y\in\mathcal{Y}}\thetaAlt^\top y \le \thetaAlt^\top y_0}$.
\end{remark}

In order to facilitate the understanding of these definitions, we present several examples of relative, absolute and \text{$\varepsilon$-explanations} obtained with \texttt{CF-OPT} for the Warcraft Maps structured learning pipeline in \cref{app:sec:examples}.

\subsection{The Explanation Problem}
\label{sec:expl_problem}
Several desirable properties have been identified for counterfactual explanations \citep{Wachter2017, Verma2020}. Arguably, the two most important ones are proximity and plausibility. The first property ensures that the counterfactual is close to the initial context, and the latter ensures that it is close to the data manifold.

We introduce the function $h$ to measure the satisfaction of the explanation criterion, that is:
\begin{equation*}
    h(x) = 
    \begin{cases}
        \varphi(x)^\top\left(\yAlt-y_0\right) \textrm{for relative explanations,}\\
        \varphi(x)^\top\left(\yAlt-y^*(x)\right) \textrm{for absolute explanations,}\\
        \varphi(x)^\top\left(\left(1+\varepsilon\right)y^*(x)-y_0\right) \textrm{for $\varepsilon$-explanations.}
    \end{cases}
\end{equation*}
A context $x$ is a valid explanation if $h(x) \le 0$. Without loss of generality, we consider thereafter the explanation constraint to be an equality, that is $h(x)=0$, which is always possible by introducing a slack variable.

\textbf{Constrained optimization.} Let $\ell$ be a differentiable function used to measure proximity in the contextual feature space $\mathcal{X}$, e.g., the squared Euclidean distance. A close and plausible counterfactual explanation $\xAlt$ is a solution to the constrained optimization problem given by:

\vspace{-0.7cm}
\begin{subequations}
    \label{opt:exp}
    \begin{align}
        \min_{x \in \mathcal{X}} \quad & \ell(x_0,\,x) \\
        \subto \quad & h(x) = 0, \\
        & x \in \mathcal{D}
    \end{align}
\end{subequations}
\vspace{-0.7cm}

where $\mathcal{D}$ is a \emph{region of plausibility}.

\paragraph{Region of plausibility}
The counterfactual explanation literature has not converged on a definition of plausibility. We outline the following
desirable properties for such a region. 
\begin{inlinelist}
    \item A region $\mathcal{D}$ that allows searching for close explanations by having small expected distance to initial contexts, i.e., $\mathbb{E}_{x_0\sim\mathbb{P}_\text{data}}\left[\min_{x \in \mathcal{D}} \ell(x_0,\,x)\right]$ is small;
    \item A low volume  $\textrm{Leb}(\mathcal{D})$ w.r.t the Lebesgue measure in order to avoid the presence of low probability areas within $\mathcal{D}$;
    \item Respecting the symmetries of the problem: e.g, if $\mathbb{P}_\textrm{data}$ is isotropic, $\mathcal{D}$ should also be.
\end{inlinelist}
The constraint $x\in\mathcal{D}$ therefore models the plausibility of the explanation, and is implicit in the sense that the distribution $\mathbb{P}_\textrm{data}$ is unknown. However, it is critical to ensure that the explanation is not an adversarial example as illustrated in \cref{fig:adv}. We provide a method to build such a region in the next section.

The link between adversarial examples and counterfactual explanations is well documented in the predictive setting, where one aims to explain classifiers \citep{browne_semantics_2020, pawelczyk_exploring_2021, freiesleben_intriguing_2022}. Both consist in finding the closest input that flips a model's prediction and share the same optimization formulation. Our setting is more complex since the model to be explained is a structured learning pipeline and not a simple classifier. Still, there are clear connections between explanations of decisions and adversarial examples. Absolute explanations present a constrained optimization formulation close to that of  white-box targeted adversarial attacks, that aim at finding the closest input that makes the output probability of a given class the highest one \citep{ma_understanding_2021}. Likewise, \text{$\varepsilon$-explanations} are similar to untargeted adversarial attacks.
\vspace{-2mm}
\begin{figure*}
    \centering
    \includegraphics[width=0.99\textwidth]{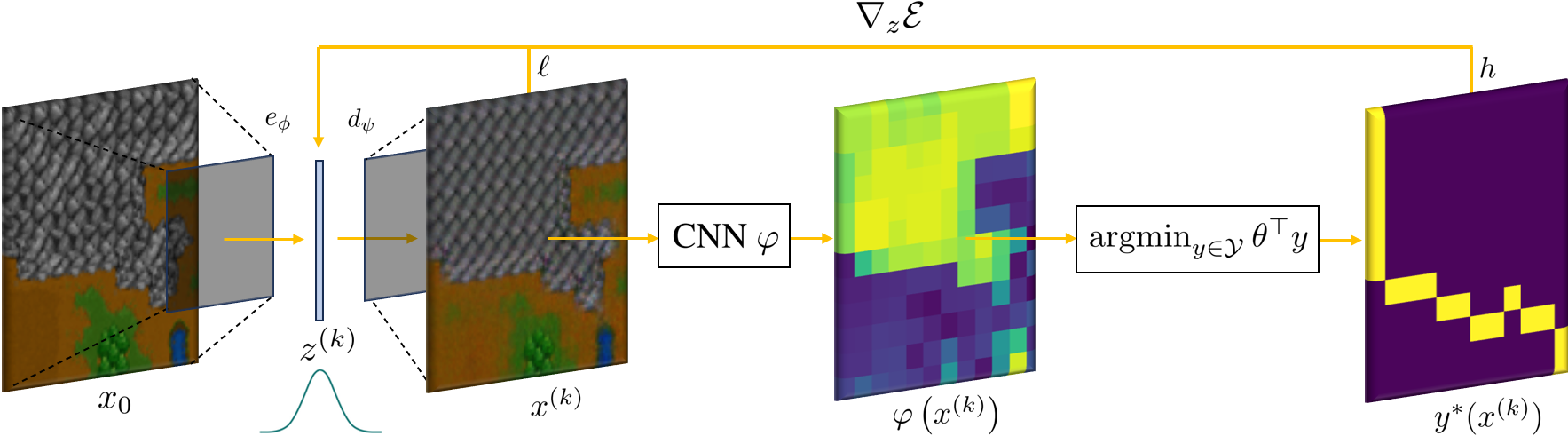}
    \vspace{-2mm}
    \caption{Pipeline of \model{CF-OPT} for plausible explanations in high-dimensional spaces. The input $x$ is encoded and decoded using a CA-VAE. The reconstructed context $\tilde{x}$ is given to the prediction model $\varphi$, a CNN in our guiding example, to obtain the parameters $\theta$. The parameterized optimization model is finally solved to obtain the decision $y^*$.}
    \label{fig:cfopt_pipeline}
    \vspace{-4mm}
\end{figure*}

\section{Modeling Plausibility with a Variational Autoencoder}
In \model{CF-OPT}, we use a VAE to get a meaningful and tractable formulation of the plausibility constraint $x \in \mathcal{D}$.
More precisely, 
\begin{inlinelist}
    \item we reformulate the plausibility constraint in the latent space, and
    \item we relax it into a soft constraint via a regularized objective. This strategy is made even more efficient by
    \item tailoring the VAE learning algorithm to obtain a model better adapted to modeling plausibility in our structured setting.
\end{inlinelist}

\paragraph{Background.} VAEs are probabilistic generative models made of two components. First, an encoder $e_\phi$ parameterized by $\phi$ maps any input $x\in\mathbb{R}^{n_x}$ to a distribution $e_\phi(\mathbf{z}|x)$ over the latent space $ \mathcal{Z} \subseteq \mathbb{R}^{n_z}$, with $n_z \ll n_x$. Then, a decoder $d_\psi$ parameterized by $\psi$ maps any latent variable $z\in\mathbb{R}^{n_z}$ back to the feature space in the form of a distribution $d_\psi(\mathbf{x}|z)$. 
The interest of a VAE is to provide a low-dimensional approximation of the unknown manifold underlying the data.
The encoder $e_\phi$ allows for representing $x$ in a much lower dimensional latent variable $z$, and the decoder $d_\psi$ tries to reconstruct $x$ from $z$.
We will note thereafter $e_\phi(x)=\Exp \left[e_\phi(\mathbf{z}\vert x)\right]$ and $d_\psi(z)=\Exp \left[d_\psi(\mathbf{x}\vert z)\right]$.

Training a VAE is usually done by maximizing the following lower bound on its log-likelihood, called ELBO (for Evidence Lower Bound, \citet{kingma_vae2022}):
\begin{equation}
    \label{eq:ELBO}
    \begin{split}
    \mathcal{L}(\phi, \psi; x_i)= \Exp_{z\sim e_\phi(\mathbf{z}|x_i)}\left[\log d_\psi(x_i|z)\right] \\
    -\DKL\left(e_\phi(\mathbf{z}|x_i)||p(\mathbf{z})\right)
    \end{split}
\end{equation}
where $x_i$ is a sample from the training set, $\DKL$ is the Kullback-Leibler~(KL) divergence, and $p(\mathbf{z})$ is a prior distribution over the latent variable. As is often done, we set the latter to be an isotropic, centered, multivariate Gaussian distribution: $p(\mathbf{z})= \mathcal{N}\left(\mathbf{z};\, 0,\, \mathbf{I}_{n_z}\right)$.

\subsection{Latent Space Counterfactual Search}

The VAE, once trained on the same dataset $\varphi$ was trained on, allows us to benefit from an approximation of the data manifold by mapping its latent space $\mathcal{Z}$ back to feature space via $d_\psi$. Thus, we reformulate the plausibility constraint $x\in \mathcal{D}$ of problem \eqref{opt:exp} in latent space, and state that a close and plausible ---latent--- counterfactual explanation $\zAlt$ is a solution to the problem:

\vspace{-0.7cm}
\begin{subequations}
    \label{opt:exp_z}
    \begin{align}
        \min_{z \in \mathcal{Z}} \quad & \ell\left(x_0,d_\psi(z)\right) \label{exp_z_obj}\\
        \subto \quad & \chi (z) = 0, \\
        \label{exp_z_pl_const}& z \in \mathcal{D}_\mathcal{Z}
    \end{align}
\end{subequations}
\vspace{-0.7cm}

where $\chi(z)=h\circ d_\psi(z)$, and $\mathcal{D}_\mathcal{Z}$ is a region of plausibility in latent space. We recover an explanation by taking \text{$\xAlt=d_\psi(\zAlt)$}.

\begin{remark}
In several applications, the feature space $\mathcal{X}$ might not be endowed with a meaningful metric to measure the proximity of a tentative counterfactual $\xAlt$ w.r.t $x_0$. 
A convenient substitute is then to measure this distance in the latent space, and use \text{$\ell(x_0, \zAlt) = \vert\vert e_\phi(x_0) - \zAlt \vert\vert_2^2$} as objective \citep{deudon_learning_2018}.

\end{remark}

\subsection{$\mathcal{D}_{\mathcal{Z}}$ as a Latent Hypersphere}
\label{sec:plausib_reg}

It turns out that defining a region of plausibility is easier in the latent space.
Indeed, we suggest using $\mathcal{D}_{\mathcal{Z}}$ as the thickened hypersphere
$$\mathcal{D}_{\mathcal{Z}} = \big\{z\in\mathcal{Z}\colon \|z\|_2 \in [C_{n_z} - \kappa ,C_{n_z}+\kappa]\}$$
with $C_{n_z}=\Exp_{z\sim p(\mathbf{z})} \vert\vert z \vert\vert_2\approx \sqrt{n_z}$ and $\kappa > 0$ a small constant.
Practically, to ease computations and avoid potential infeasible problems, we use a soft version of the constraint $z \in \mathcal{D}_\mathcal{Z}$, leading to the counterfactual explanation problem

\vspace{-0.7cm}
\begin{subequations}
    \label{opt:exp_z_reg}
    \begin{align}
        \min_{z \in \mathcal{Z}} \quad & \ell\left(x_0,d_\psi(z)\right)+ \Omega(z) \label{exp_z_obj_reg}\\
        \subto \quad & \chi (z) = 0,
    \end{align}
\end{subequations}
\vspace{-0.7cm}

with \text{$\Omega(z)=\beta\left(\vert\vert z\vert\vert_2 - C_{n_z}\right)^2$} being our proposed plausibility regularization, and $\beta>0$ a hyperparameter controlling its weight.

To understand our definition of $\mathcal{D}_\mathcal{Z}$, recall that since the KL divergence in \cref{eq:ELBO} forces the distribution of latent codes to match that of the prior $p(\mathbf{z})$, a standard Gaussian is a good approximation of the push-forward of $\mathbb{P}_{\textrm{data}}$ through the encoder $e_\phi$. In \cref{sec:prior_approx}, we provide numerical justifications for this approximation. The Gaussian being isotropic, it is natural to define $\mathcal{D}_\mathcal{Z}$ as \text{$\{z\in\mathcal{Z} \colon a \leq \vert\vert z\vert\vert_2\leq b\}$}. Besides, it is well known that the $\ell^2$ norm of a high-dimensional standard Gaussian concentrates around its expectation \citep{biau_high_dimensional_2013}, leading to our definition of $\mathcal{D}_\mathcal{Z}$.

In \cref{sec:hyper_proof}, we give theoretical insights into why this region is optimal w.r.t the desirable properties we identified in \cref{sec:expl_problem}. Moreover, as a benchmark, we experimentally show in \cref{sec:impact_plausib_reg} that using this hypersphere instead of a centered ball leads to improved performance (the corresponding region $\Tilde{\mathcal{D}}_\mathcal{Z}$ being of the form \text{$\{z\in\mathcal{Z} \colon p(z) \geq \delta\}$} and the corresponding regularization being \text{$\Tilde{\Omega}(z)=\beta\vert\vert z\vert\vert_2^2$}).

\subsection{Tailoring VAE Training with a Cost-Aware Loss}

The costs predicted after VAE-reconstruction $\varphi(\tilde{x})$ can be quite far from $\varphi(x)$, as training a VAE to maximize the traditional ELBO is agnostic to the downstream task in our specific setting.
This is unfortunate since a poor accuracy of the reconstructed costs reduces the quality of the obtained counterfactuals: indeed, it can lead to perturbations in the pipeline's output associated to the initial context $x_0$.

We adopt an empirical approach to this issue and penalize the distance between predicted costs before and after reconstruction during training. More precisely, we suggest using the following term for sample $x_i$ in the maximized objective when training our VAE in a cost-aware fashion:
\begin{equation}
    \label{eq:CAVAEobj}
    \begin{split}
    \mathcal{L}_{\text{CA}} (\phi, \psi; &  x_i) = \mathcal{L}(\phi, \psi; x_i) \\
    - & \alpha \, \Exp_{z\sim e_\phi (\mathbf{z}\vert x_i)} \left[ \vert\vert\varphi(x_i) - \varphi\left(d_\psi(z)\right) \vert\vert_2^2 \right],
    \end{split}
\end{equation}
where $\mathcal{L}(\phi, \psi; x_i)$ is the ELBO traditionally used, and \text{$\alpha>0$} is a hyperparameter controlling the regularization weight. Thus, this learning objective takes into account the \textit{cost-reconstruction error}, in addition to the traditional \textit{feature-reconstruction error} (the expectation term in \cref{eq:ELBO}). During training, this additional term is approximated using a Monte-Carlo estimate, as is already usually done when approximating the feature-reconstruction loss (naturally, we use the same samples to compute both).

\begin{remark}
    We did not observe a significant impact of the dimension of the latent space on the method, other than its known impact on the ability of the VAE to generate high-quality samples. Hence, the architectural choices follow the usual concerns when training a VAE.
\end{remark}

\section{Computing Counterfactuals}
In general, the explanation problem, whether formulated in feature space or latent space, is non-convex and has no closed-form solution. Our algorithm to obtain explanations is based on constrained differential optimization and, in particular, the modified differential method of multipliers~(MDMM) of \citet{Platt1987constrained}.

Similarly to most works on end-to-end training, we also need to propagate a gradient through a discrete optimization layer. However, there is an important difference: the pipeline is already trained. Thus, we compute a gradient w.r.t the covariate's features and not the model's parameters as is done when training the pipeline.

\subsection{Augmented Lagrangian and its Gradient}
The MDMM is based on the following relaxation of the constrained optimization problem \ref{opt:exp_z_reg}, called an \textit{energy function} by \citet{Platt1987constrained}, and commonly known as the augmented Lagrangian \citep{Boyd2011distributed}:
\begin{equation}
    \label{auglag}
    \mathcal{E}(z, \lambda) = \ell \left(x_0, d_\psi(z) \right) + \Omega(z) + \lambda \, \chi \left(z \right) + \frac{\rho}{2}\left(\chi \left( z \right) \right)^2
\end{equation}

where $\lambda \in \mathbb{R}$ is the dual variable of the explanation constraint and $\rho > 0$ is a constant hyperparameter, called the damping term.

The idea of the MDMM is to perform a gradient descent on~$\mathcal{E}$ w.r.t $z$ and a gradient ascent on~$\mathcal{E}$ w.r.t $\lambda$ in order to reach a local minimum $z^*$ that satisfies the constraint $\chi(z^*) = 0$.

\begin{remark}
    The primal variable $z$ is initialized at \text{$z^{(1)} = e_\phi(x_0)$}, which is the only  time the encoder $e_\phi$ is used per explanation task.
\end{remark} This algorithm is closely related to another first-order constrained optimization algorithm: the method of multipliers~(MoM) \citep{Boyd2011distributed}. The main difference is that, at each iteration, the optimization problem is solved to optimality instead of updating $z$ with a gradient descent step. The MDMM is more convenient in our setting since solving $\min_z \mathcal{E}(z, \lambda)$ for each $\lambda$ is difficult.

As it involves the composition of complex deep architectures (the decoder and the prediction model), the optimization problem considered is highly non-convex. The MDMM can therefore only provide convergence to a local minimum, under rather technical assumptions on its initialization which need not be detailed here \citep{Platt1987constrained}. However, our numerical experiments highlight that it is very efficient at obtaining feasible and close solutions.
 
\textbf{Gradients.} 
We note $x = d_\psi(z)$. The energy function~$\mathcal{E}$ is differentiable almost everywhere w.r.t $z$ (everywhere in the case of relative explanations, the problem coming from differentiating $y^*(x)$ when $\argmin_{y\in\mathcal{Y}}\varphi(x)^\top y$ contains more than one element). Moreover, we have:
\vspace{-0.1cm}
\begin{equation}
    \label{eq:expl_grad}
    \nabla \chi(z) = 
    \begin{cases}
        \begin{aligned}
            &J_{\varphi \circ d_\psi}(z)\left(\yAlt-y_0\right) & \hfill & \text{(relative exp.)}\\
            &J_{\varphi \circ d_\psi}(z)\left(\yAlt-y^*(x)\right) & \hfill & \text{(absolute exp.)}\\
            &J_{\varphi \circ d_\psi}(z)\left(\left(1+\varepsilon\right)y^*(x)-y_0\right) & \hfill & \text{($\varepsilon$-exp.)}
        \end{aligned}
    \end{cases}
\end{equation}
These gradients can be derived easily since \text{$\nabla_x y^*(x) = 0$} almost everywhere due to $y^*$ being piecewise constant, as the solution of a linear optimization problem. The Jacobian $J_{\varphi \circ d_\psi}(z)$ is computed via automatic differentiation through decoder $d_\psi$ and prediction model $\varphi$.

\begin{remark}
    If $\;\min_{y\in\mathcal{Y}}\varphi(x)^\top y\;$ is negative, we only have to change $\,(1+\varepsilon)\,$ to $\,(1-\varepsilon)\,$ inside the expression of the gradient for $\varepsilon$-explanations (see \cref{positive_cost}).
\end{remark}

\subsection{Our Algorithm}
Our algorithm is given in \cref{alg:cf_opt_latent} in \cref{app:algo}. The parameters $K$ and $c_{\max}$ are hyperparameters that define the maximum number of iterations and the maximum number of non-improving iterations. We define an iteration $k$ as improving if $x^{(k)}$ satisfies the explanation criterion, and if $\ell (x_0, x^{(k)}) + \Omega(z^{(k)}) \leq u \left( \ell (x_0, x^{(j)})+ \Omega(z^{(j)})\right)$, where $j$ is the index of the last improving iteration and $u\in \left] 0, 1\right[$ is an update tolerance hyperparameter.

If the plausibility of the explanation is not taken into account, the use of a VAE is not necessary and the feature space formulation of \texttt{CF-OPT} is given in \cref{alg:cf_opt_feature}.

The computational effort of finding an explanation of a decision lies in computing the gradient in \cref{eq:expl_grad}. For absolute and \text{$\varepsilon$-explanations}, the linear optimization problem $\min_{y\in\mathcal{Y}}\varphi(x^{(k)})^\top y$ is solved once per iteration. The solution $y^*(x^{(k)})$ is used twice at each iteration: to check whether the explanation criterion is satisfied and to compute the gradient~$\nabla_z \mathcal{E}$. Relative explanations do not require to solve this problem and are thus cheaper to compute.

\section{Numerical Study}
\label{sec:numerical}
In this section, we evaluate the ability of \model{CF-OPT} to obtain plausible explanations of decisions in a high-dimensional setting. We focus on the problem of finding shortest paths on Warcraft maps, which has been our guiding example throughout the paper. Warcraft maps are $96\times 96$ RGB images with continuous color values. Every area is walkable, and crossing them results in different costs. The true, unknown costs can only take discrete values (associated with the terrain being a forest, rock, water, etc.), but the Convolutional Neural Network (CNN) $\varphi$ used to predict them from the maps outputs continuous ones. The output of the full structured learning pipeline is the shortest path from the upper-left corner to the lower-right one.

Our experiments study the value of using the hypersphere regularization and training a VAE in a cost-aware fashion. We provide the details of our experimental setting, the architecture of our models, as well as other results with low-dimensional tabular data in \cref{app:numeric}.

Our experiments are implemented in Python and run on four cores of an Intel Core i7-8565U CPU @ 1.80GHz and use 16GB RAM. The code used to generate all the results in this paper is available publicly at \url{https://github.com/GermainVivierArdisson/CF-OPT} under an MIT license.

\subsection{Experimental Setting}
\label{sec:exp_warcraft}
We follow the experimental setting of \citet{vlastelica2019diff} and used subsequently in several works \citep{Dalle2022, tang2022pyepo, McKenzie2023faster}. The training set $\left\lbrace x_i, \theta_i, y_i \right\rbrace_{i=1}^{N}$ is made of $N=10 000$ examples of Warcraft maps as well as their associated true costs and shortest paths.

\textbf{Relative regret metric.} We introduce here the relative regret metric, which is used in our experiments to quantify the impact of both the cost-aware and plausibility regularization. The relative regret $\reg(y, \theta)$ of a feasible solution $y\in \mathcal{Y}$ for a cost $\theta$ is defined as:
\begin{equation}
    \reg(y, \theta) = \frac{\theta^\top y - \theta^\top y^*(\theta)}{\vert\theta^\top y^*(\theta)\vert},
\end{equation}
where $\theta$ is a cost vector and and  $y^*(\theta) = \argmin_{y\in\mathcal{Y}}\theta^\top y$ is the optimal solution associated to $\theta$.

\textbf{VAE models.} All VAEs are implemented as CNNs with convolutions for the encoder and transposed convolutions for the decoder. The default latent space dimension is taken to be $64$.

\textbf{Selection of explanation tasks.} In our experiments, we repeatedly solve explanation problems using initial maps and alternative paths to obtain $N_\textrm{exp}$ explanations. The initial maps are all different and never seen during the training of any VAE. Further, to avoid trivial explanation tasks, the alternative paths defining explanation problems are always taken to be different from the shortest path associated to the initial map.

\subsection{Impact of Cost-Aware Training}
First, we quantify the impact of the cost-aware regularization in the training objective of the VAE in \cref{eq:CAVAEobj} by measuring the reconstruction of the optimal path after the VAE. We measure this with three metrics:
\begin{enumerate}[topsep=-0.8\parskip, itemsep=-3pt, wide=5pt, label=(\roman*)]
    \item the relative regret $\reg\left(y^*(\varphi(x_i)),\, \varphi(\Tilde{x}_i)\right)$ of the initial optimal solution associated to $x_i$ in the reconstructed context $\Tilde{x}_i$,
    \item the relative regret $\reg\left(y^*(\theta_i),\, \varphi(\Tilde{x}_i)\right)$ of the optimal path associated to the true cost $\theta_i$ in the reconstructed context $\Tilde{x}_i$,
    \item the squared $\ell^2$ norm $||x_i - \Tilde{x}_i||_2^2$, the natural metric for evaluating the feature-reconstruction performances of the VAE.
\end{enumerate}

We train several VAEs, with varying cost-aware regularization weight $\alpha$. Notice that using $\alpha = 0$ recovers the traditional cost-agnostic VAE. All VAEs are trained until convergence, with early stopping to avoid overfitting. After training, we measure the average of each performance metric over a test set of $1000$ data points. 
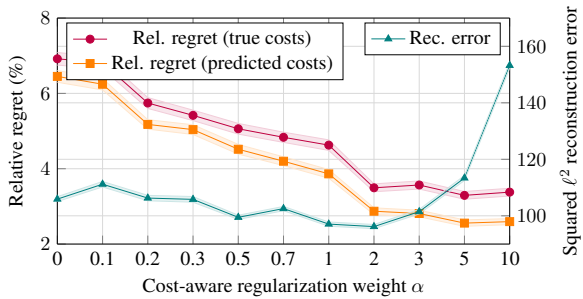
\begin{figure}[ht]
    \centering
    \vspace{-2mm}
    \resizebox{0.95\linewidth}{!}{\begin{tikzpicture}
\pgfplotsset{
    scale only axis,
    height = 4cm,
    width  = 8cm,
    xmin=0, xmax=10
}

\begin{axis}[
    axis y line*=left,
    ymin=2, ymax=8,
    xlabel= Cost-aware regularization weight $\alpha$,
    ylabel= Relative regret (\%),
    legend style={at={(0.02,0.98)},anchor=north west},
    symbolic x coords ={0, 0.1, 0.2, 0.3, 0.5, 0.7, 1, 2, 3, 5, 10},
    xtick = data
]

\addplot+[true_regretStyle] table [x index = {0}, y index = {1}, col sep=comma]{plots/csv/warcraft/relative_regret_and_reconstruction.csv};
\addplot+[name path=A, purple!20, mark=none, forget plot] table [x index = {0}, y index = {2}, col sep=comma]{plots/csv/warcraft/relative_regret_and_reconstruction.csv};
\addplot+[name path=B, purple!20, mark=none, forget plot] table [x index = {0}, y index = {3}, col sep=comma]{plots/csv/warcraft/relative_regret_and_reconstruction.csv};
\addplot[purple, fill opacity=0.1] fill between[of=A and B];


\addplot+[predicted_regretStyle]  table [x index = {0}, y index = {4}, col sep=comma]{plots/csv/warcraft/relative_regret_and_reconstruction.csv};

\addplot+[name path=A, orange!20, mark=none, forget plot] table [x index = {0}, y index = {5}, col sep=comma]{plots/csv/warcraft/relative_regret_and_reconstruction.csv};
\addplot+[name path=B, orange!20, mark=none, forget plot] table [x index = {0}, y index = {6}, col sep=comma]{plots/csv/warcraft/relative_regret_and_reconstruction.csv};
\addplot[orange, fill opacity=0.1] fill between[of=A and B];
\legend{Rel. regret (true costs), , Rel. regret (predicted costs)}

\end{axis}

\begin{axis}[
  axis y line*=right,
  axis x line=none,
  ymin=90, ymax=170,
  ylabel=Squared $\ell^2$ reconstruction error,
  symbolic x coords ={0, 0.1, 0.2, 0.3, 0.5, 0.7, 1, 2, 3, 5, 10},
  xtick = data
]
\addplot+[reconstStyle] table [x index = {0}, y index = {7}, col sep=comma]{plots/csv/warcraft/relative_regret_and_reconstruction.csv};
\addlegendentry{Rec. error}
\addplot+[name path=A, teal!20, mark=none, forget plot] table [x index = {0}, y index = {8}, col sep=comma]{plots/csv/warcraft/relative_regret_and_reconstruction.csv};
\addplot+[name path=B, teal!20, mark=none, forget plot] table [x index = {0}, y index = {9}, col sep=comma]{plots/csv/warcraft/relative_regret_and_reconstruction.csv};
\addplot[teal, fill opacity=0.1] fill between[of=A and B];

\end{axis}
\end{tikzpicture}}
    \vspace{-2mm}
    \caption{Comparison of VAE and Cost-Aware VAE for varying~$\alpha$}
    \label{fig:ca_reg}
    \vspace{-2mm}
\end{figure}

The results in \cref{fig:ca_reg} show that the relative regret metrics defined above clearly decrease when increasing the regularization weight. Hence, introducing a cost-aware regularization is very powerful to decrease cost-reconstruction error, and to stabilize the optimal paths associated with the Warcraft maps under VAE-reconstruction. As expected, the relative regret is lower when taking the initial optimal path associated with the costs predicted by $\varphi$ rather than the true ones. \cref{fig:ca_reg} further shows that the reconstruction error decreases for $0 < \alpha \le 2$. This result suggests that increasing the regularization weight even guides the training process and helps the model to reconstruct images. Consequently, we set the cost-regularization parameter to $\alpha=2$ for the rest of the experiments.

\begin{figure}[ht]
    \centering
    \vspace{-1mm}
    \resizebox{0.9\linewidth}{!}{\begin{tikzpicture}

    \pgfplotsset{
        scale only axis,
        xmin=0, xmax=10
    }
    \begin{axis}[
            height = 4cm,
            width  = 8cm,
            enlarge x limits = 0,
            enlarge y limits = 0,
            xlabel = {Cost-aware regularization weight $\alpha$},
            ylabel = {Loss of $x_{\textrm{best}}$},
            symbolic x coords ={0, 0.1, 0.2, 0.3, 0.5, 0.7, 1, 2, 3, 5, 10},
            xtick = data
        ]

        \addplot+[losscavaeStyle] table [x index = {0}, y index = {1}, col sep=comma]{plots/csv/warcraft/loss_cavae.csv};
        \addlegendentry{Loss (mean)}
        \addplot+[name path=A, violet!20, mark=none, forget plot] table [x index = {0}, y index = {2}, col sep=comma]{plots/csv/warcraft/loss_cavae.csv};
        \addplot+[name path=B, violet!20, mark=none, forget plot] table [x index = {0}, y index = {3}, col sep=comma]{plots/csv/warcraft/loss_cavae.csv};
        \addplot[violet, fill opacity=0.1, forget plot] fill between[of=A and B];
    \end{axis}
\end{tikzpicture}}
    \vspace{-2mm}
    \caption{Loss of the obtained explanations when using Cost-Aware VAE with varying $\alpha$}
    \label{fig:plot_loss_cavae}
    \vspace{-2mm}
\end{figure}
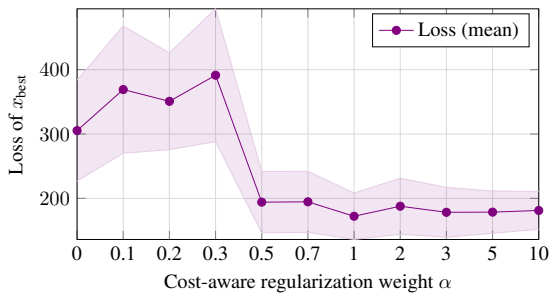

To further highlight the impact of the cost-aware regularization w.r.t the explanation problem, we analyze the loss of obtained explanations when using trained Cost-Aware VAEs with different regularization weights. For each weight, we run \model{CF-OPT} on $N_\textrm{exp}=50$ different absolute explanation tasks. We use a step size $\gamma=0.003$, maximum number of iterations $K=3000$, maximum number of non-improving iterations $c_{\max} = 50$, and corresponding update tolerance $u=0.9$. We then measure the loss of the best solution according to our best VAE using latent hypersphere plausibility regularization with $\beta = 10$, and a feature space objective. The loss is thus equal to $||x_{\rm{best}}-x_0||^2_2 + 10\left(||z_{\rm{best}}||_2-C_{64}\right)^2$. As seen in \cref{fig:plot_loss_cavae}, Cost-Aware VAEs trained with $\alpha\geq 0.5$ lead to significantly better counterfactual explanations in terms of the resulting loss.

\subsection{Impact of Plausibility Regularization} 
\label{sec:impact_plausib_reg}
We now analyze the impact of the proposed latent hypersphere regularization on the plausibility of the obtained counterfactual explanations. Instead of relying only on subjective visual comparisons of the results, we use two different metrics to quantify plausibility.

\begin{figure}[ht]
    \centering
    \vspace{-1mm}
    \resizebox{0.9\linewidth}{!}{\begin{tikzpicture}

    \pgfplotsset{
        scale only axis,
        xmin=0, xmax=100
    }
    \begin{axis}[
            height = 4cm,
            width  = 8cm,
            enlarge x limits = 0,
            enlarge y limits = 0,
            xlabel = {Plausibility regularization weight $\beta$},
            ylabel = {Reconstruction error on $x_{\textrm{best}}$},
            symbolic x coords ={0, 0.1, 0.3, 0.5, 1, 3, 5, 10, 30, 50, 100},
            xtick = data
        ]

        \addplot+[reconstStyle] table [x index = {0}, y index = {1}, col sep=comma]{plots/csv/warcraft/reconstruction_beta.csv};
        \addlegendentry{$||\tilde{x}_{\textrm{best}} - x_{\textrm{best}}||_2^2$}
        \addplot+[name path=A, teal!20, mark=none, forget plot] table [x index = {0}, y index = {2}, col sep=comma]{plots/csv/warcraft/reconstruction_beta.csv};
        \addplot+[name path=B, teal!20, mark=none, forget plot] table [x index = {0}, y index = {3}, col sep=comma]{plots/csv/warcraft/reconstruction_beta.csv};
        \addplot[teal, fill opacity=0.1, forget plot] fill between[of=A and B];

    \end{axis}
\end{tikzpicture}}
    \vspace{-2mm}
    \caption{Reconstruction error of a Cost-Aware VAE ($\alpha=2$) on the best solution obtained $x_{\textrm{best}}$, with varying latent hypersphere regularization weight $\beta$.}
    \label{fig:plot_reconstruction_beta}
    \vspace{-2mm}
\end{figure}
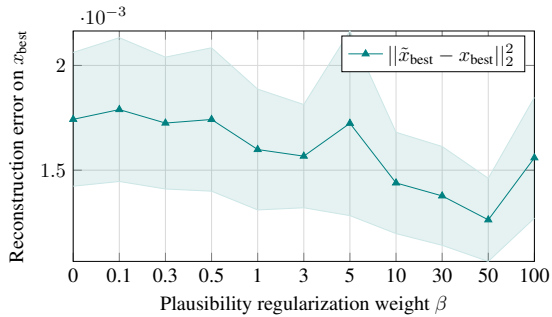

\textbf{Reconstruction error.} First, we use a well-known metric in outlier detection \citep{An2015VariationalAB} measuring the reconstruction error of a VAE trained on the corresponding dataset when evaluated on the potential outlier. We select $N_\textrm{exp}=40$ absolute explanation tasks as above and run \model{CF-OPT} with the same hyperparameters but varying the weight $\beta$ of the latent hypersphere regularization. Then, we measure the reconstruction error with a Cost-Aware VAE trained on the Warcraft maps dataset with $\alpha=2$, when evaluated on the returned best solutions $x_{\textrm{best}}$. The results are displayed in \cref{fig:plot_reconstruction_beta} and show that the proposed latent hypersphere regularization leads to more plausible explanations w.r.t this outlier detection metric.

\textbf{Decision-focused reconstruction error.} Our second metric is specifically designed for our setting. Instead of simply measuring the reconstruction error in the feature space as above, we measure the stability of the optimal path associated to the returned best solution $x_{\textrm{best}}$ under reconstruction with a Cost-Aware VAE. This metric is more suited to detect adversarial explanations since adversarial perturbations tend to simply vanish when passing them through a VAE, due to the information compression aspect of the model (which then acts as a denoising model), thus leading to a small reconstruction error. To do so, we measure the relative regret $\reg\left(y^*(\varphi(x_{\textrm{best}})),\,\varphi(\Tilde{x}_{\textrm{best}})\right)$ of the shortest path associated with $x_{\textrm{best}}$ in the reconstructed context $\Tilde{x}_{\textrm{best}}$ when using a Cost-Aware VAE trained with $\alpha=2$. This metric extends the idea of reconstruction error to the entire data-driven optimization pipeline.
\begin{figure}[ht!]
    \centering
    \vspace{-1mm}
    \begin{subfigure}[t]{0.49\columnwidth}
        \includegraphics[width=\linewidth]{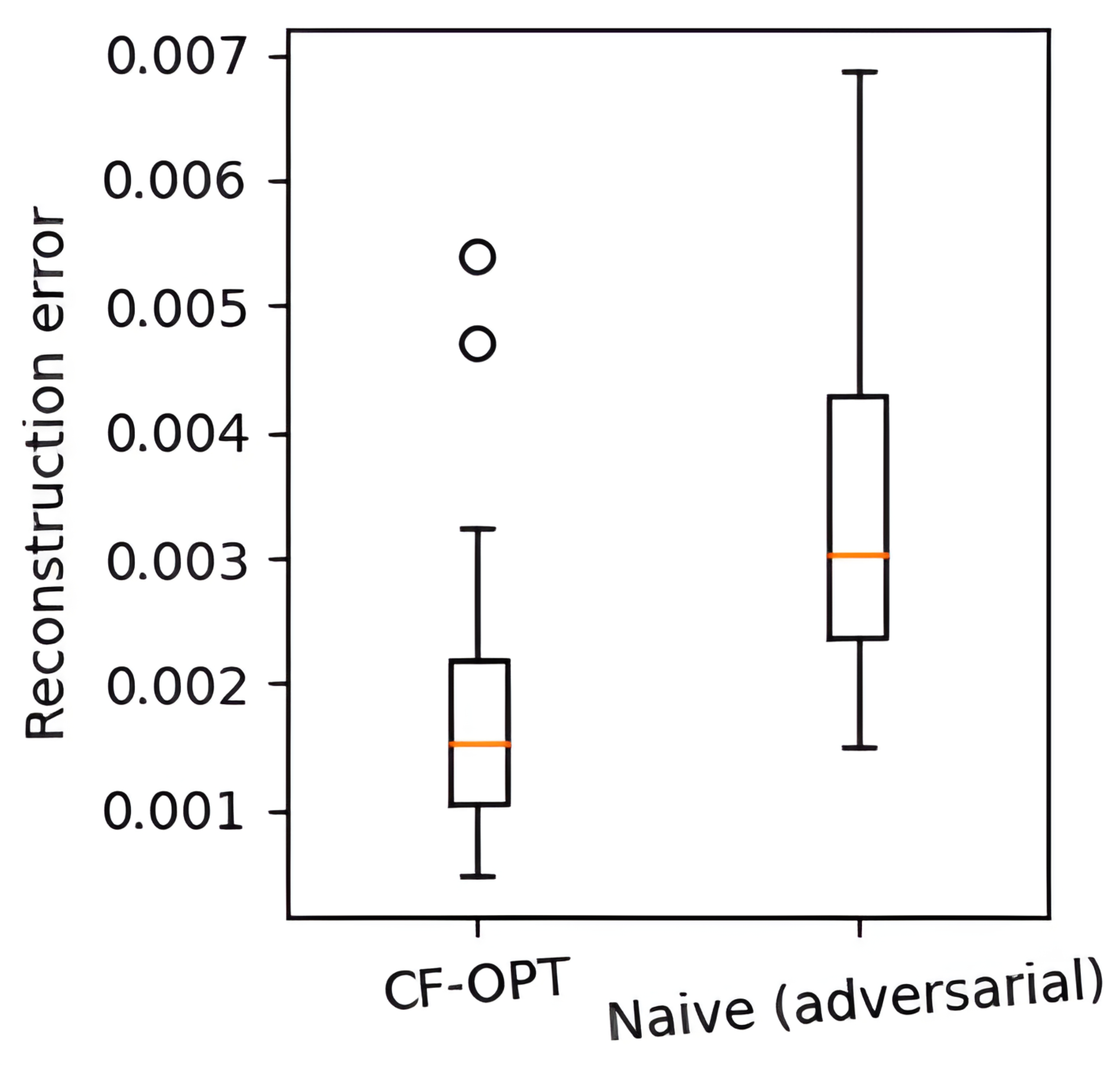}
        \caption{Reconstruction error}
    \end{subfigure}
    \hfill
    \begin{subfigure}[t]{0.49\columnwidth}
        \includegraphics[width=\linewidth]{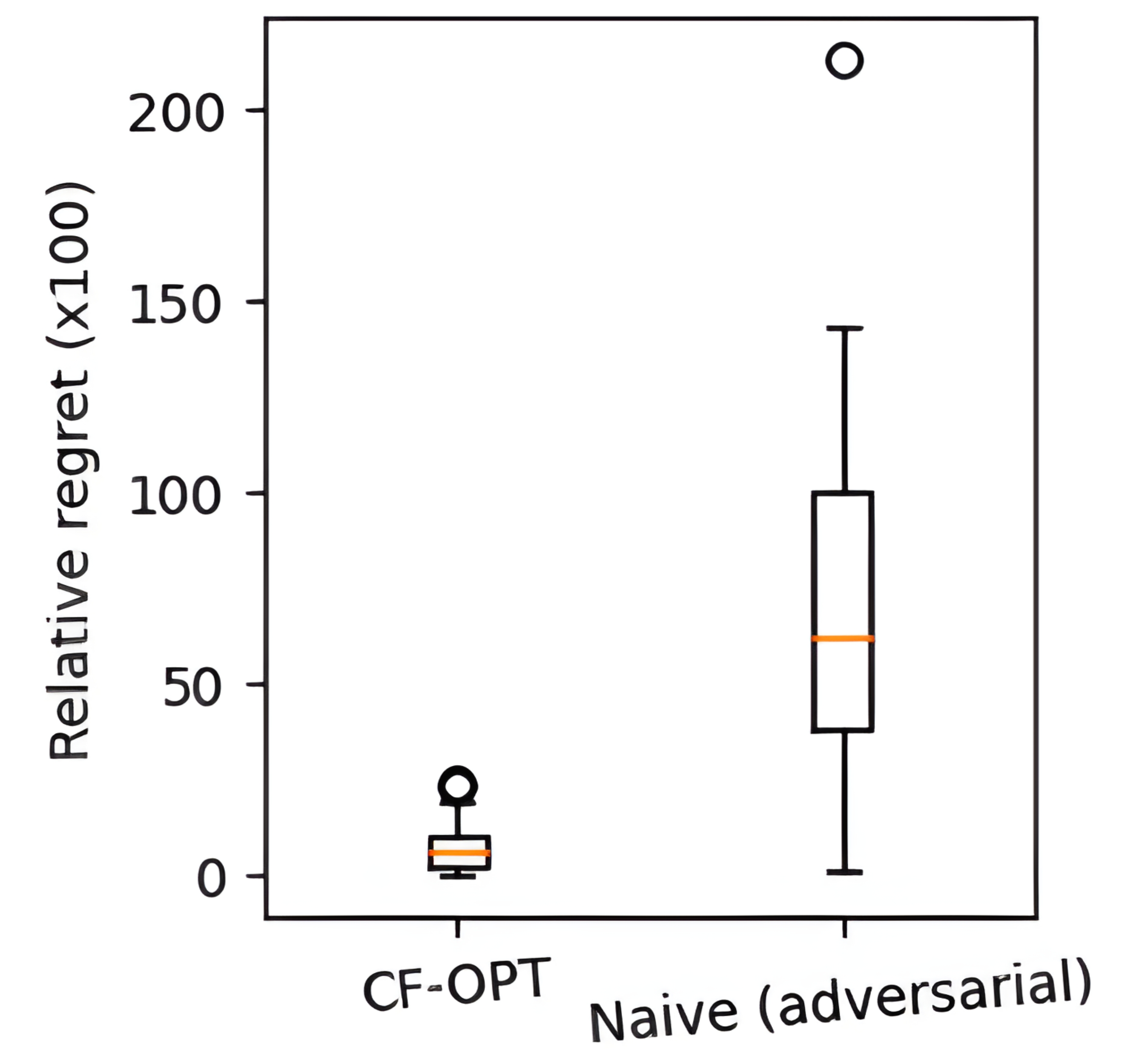}
        \caption{Decision-focused reconstruction error}
    \end{subfigure}
    \vspace{-2mm}
    \caption{(a)~Task-agnostic reconstruction error and (b)~decision-focused reconstruction error for detecting adversarial explanations.}
    \label{fig:boxplot_adv}
    \vspace{-2mm}
\end{figure}

We obtain $N_\textrm{exp}=40$ absolute explanations using a latent-space search with \model{CF-OPT}, and $N_\textrm{exp}=40$ other counterfactuals using a naive search (i.e., adversarial explanations as in \cref{fig:adv}). For \model{CF-OPT}, the VAE used for optimization is not cost-regularized ($\alpha=0$) in order to avoid biases since the proposed metric is computed using a Cost-Aware VAE itself. As shown in \cref{fig:boxplot_adv}, our decision-focused metric is more efficient at differentiating between adversarial and plausible explanations than the classical reconstruction error.

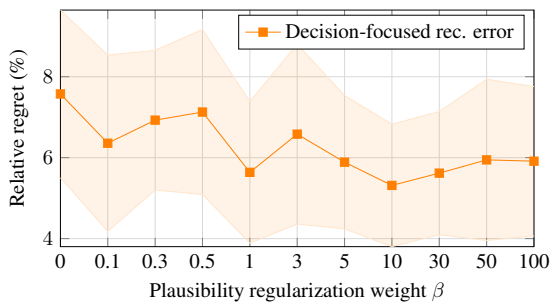
\begin{figure}[ht]
    \centering
    \resizebox{0.9\linewidth}{!}{\begin{tikzpicture}

    \pgfplotsset{
        scale only axis,
        xmin=0, xmax=100
    }
    \begin{axis}[
            height = 4cm,
            width  = 8cm,
            enlarge x limits = 0,
            enlarge y limits = 0,
            xlabel = {Plausibility regularization weight $\beta$},
            ylabel = {Relative regret (\%)},
            symbolic x coords ={0, 0.1, 0.3, 0.5, 1, 3, 5, 10, 30, 50, 100},
            xtick = data
        ]

        \addplot+[predicted_regretStyle] table [x index = {0}, y index = {1}, col sep=comma]{plots/csv/warcraft/stability_beta.csv};
        \addlegendentry{Decision-focused rec. error}
        \addplot+[name path=A, orange!20, mark=none, forget plot] table [x index = {0}, y index = {2}, col sep=comma]{plots/csv/warcraft/stability_beta.csv};
        \addplot+[name path=B, orange!20, mark=none, forget plot] table [x index = {0}, y index = {3}, col sep=comma]{plots/csv/warcraft/stability_beta.csv};
        \addplot[orange, fill opacity=0.1, forget plot] fill between[of=A and B];

    \end{axis}
\end{tikzpicture}}
    \vspace{-2mm}
    \caption{Decision-focused reconstruction error measured with an $\alpha=2$ Cost-Aware VAE, for varying latent hypersphere regularization weight $\beta$.}
    \label{fig:plot_stability_beta}
    \vspace{-2mm}
\end{figure}

Hence, we use this metric to measure the plausibility of explanations obtained with \model{CF-OPT} and latent hypersphere regularization for varying regularization weight $\beta$. Notice again that $\beta=0$ recovers the unregularized explanation objective. Similarly to the experiment shown in \cref{fig:boxplot_adv}, the decision-focused reconstruction error is computed on $N_\textrm{exp}=40$ absolute explanation tasks. Our results in \cref{fig:plot_stability_beta} show that the latent hypersphere regularization also improves the reconstruction of the pipeline's output associated with the produced counterfactuals.

\textbf{Comparison of latent hypersphere and log-likelihood regularizations.} We now investigate the value of the proposed latent hypersphere regularization $\Omega$, compared to the log-likelihood regularization $\Tilde{\Omega}$ (both introduced in \cref{sec:plausib_reg}). We select $N_\textrm{exp}=40$ absolute explanation tasks and run \model{CF-OPT} on each explanation task with a Cost-Aware VAE trained with $\alpha=2$. We measure the loss of the best solution found with latent hypersphere regularization and with log-likelihood regularization for varying regularization weight $\beta$.

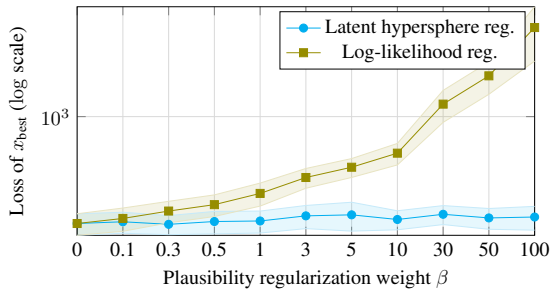
\begin{figure}[ht]
    \centering
    \resizebox{0.9\linewidth}{!}{\begin{tikzpicture}

    \pgfplotsset{
        scale only axis,
        xmin=0, xmax=100
    }
    \begin{axis}[
            ymode = log,
            height = 4cm,
            width  = 8cm,
            enlarge x limits = 0,
            enlarge y limits = 0,
            xlabel = {Plausibility regularization weight $\beta$},
            ylabel = {Loss of $x_{\textrm{best}}$ ($\log$ scale)},
            symbolic x coords ={0, 0.1, 0.3, 0.5, 1, 3, 5, 10, 30, 50, 100},
            xtick = data
        ]

        \addplot+[losssphereStyle] table [x index = {0}, y index = {1}, col sep=comma]{plots/csv/warcraft/loss_beta.csv};
        \addlegendentry{Latent hypersphere reg.}
        \addplot+[name path=A, cyan!20, mark=none, forget plot] table [x index = {0}, y index = {2}, col sep=comma]{plots/csv/warcraft/loss_beta.csv};
        \addplot+[name path=B, cyan!20, mark=none, forget plot] table [x index = {0}, y index = {3}, col sep=comma]{plots/csv/warcraft/loss_beta.csv};
        \addplot[cyan, fill opacity=0.1, forget plot] fill between[of=A and B];

        \addplot+[lossloglkStyle] table [x index = {0}, y index = {4}, col sep=comma]{plots/csv/warcraft/loss_beta.csv};
        \addlegendentry{Log-likelihood reg.}
        \addplot+[name path=A, olive!20, mark=none, forget plot] table [x index = {0}, y index = {5}, col sep=comma]{plots/csv/warcraft/loss_beta.csv};
        \addplot+[name path=B, olive!20, mark=none, forget plot] table [x index = {0}, y index = {6}, col sep=comma]{plots/csv/warcraft/loss_beta.csv};
        \addplot[olive, fill opacity=0.1, forget plot] fill between[of=A and B];

    \end{axis}
\end{tikzpicture}}
    \vspace{-2mm}
    \caption{Loss of the obtained explanations with latent hypersphere ($\Omega$) and log-likelihood ($\Tilde{\Omega}$) regularizations, with varying weight $\beta$.}
    \label{fig:plot_loss_beta}
    \vspace{-2mm}
\end{figure}

The proximity metric used is the squared euclidean distance in feature space, so that the corresponding loss is $||x_{\rm{best}}-x_0||^2_2 + \beta\left(||z_{\rm{best}}||_2-C_{64}\right)^2$ for the latent hypersphere regularization, and $||x_{\rm{best}}-x_0||^2_2 + \beta||z_{\rm{best}}||_2^2$ for the log-likelihood regularization. As can be seen in \cref{fig:plot_loss_beta}, the latent hypersphere regularization does not significantly impact the loss of the obtained counterfactual, contrary to the log-likelihood regularization which leads to an exponential increase in the loss as $\beta$ increases.

\section{Related Literature}
This work lies at the intersection of end-to-end learning, generative models, and explainable machine learning. We briefly outline some relevant literature.

\paragraph{End-to-end learning and optimization.} Integrating optimization into learning pipelines has received a lot of attention. The main challenge is to train the pipeline in an end-to-end fashion, that is, to propagate the loss associated to the objective function of the pipeline's optimization component back to its predictive component. This requires differentiating through the pipeline, particularly through the $\argmin$ operation of the optimization component. Several approaches have been proposed based, for instance, on implicit differentiation \citep{Amos2017optnet, Donti2017}, using a convex surrogate loss \citep{Elmachtoub2022}, or perturbed optimizers \citep{Berthet2020, Dalle2022}. We refer to the survey of \citet{Mandi2023decision} and \citet{Sadana2023survey} for further details.

\paragraph{Counterfactual explanations.} Counterfactual explanations have received significant attention to explain learning models \citep{Wachter2017, Verma2020}, especially in the classification setting, where they translate as the smallest alteration in the input features that changes the output's label into another, predefined one.

\paragraph{Generative models and plausible explanations.}
Generative models are also used to provide counterfactual explanations that adhere to the data manifold in the predictive setting, that is when the model to be explained is a classifier. \citet{pawelczyk_learning_2020} use a modified version of the VAE, dubbed C-CHVAE, and randomly sample latent codes on a sphere centered at the latent embedding of the initial context until the explanation condition is met. This would be very algorithmically inefficient in our setting because of the combinatorial aspect of the problem: there tends to be an exponential number of (feasible) decisions even for simpler linear problems such as the shortest path, compared to the reasonable number of classes in the predictive setting. This also prevents using methods that train separate VAE models for each class, such as the one presented by \citet{barr_counterfactual_2021}. \citet{GenCF2019} minimize a relaxed objective inside the latent space of a trained generative adversarial network (GAN) using gradient descent. Since GANs are implicit likelihood models, one can not derive a regularization on the plausibility of the generated counterfactual as easily as we did with a VAE in \cref{sec:plausib_reg}.
\section{Conclusions}
\label{sec:conclusion}
This paper presented an approach to obtain counterfactual explanations of the decisions of end-to-end optimization pipelines. It focuses on obtaining plausible explanations in high-dimensional settings such as when the input are images. One limitation of the current approach is that the covariates $x$ must be made of continuous features. Extending the work to categorical or discrete features is a promising research direction. A central component of the method is the use of a VAE, which are especially relevant in our setting because of the smaller dimension of the latent space. Another interesting direction for future work is to investigate the use of other generative models such as diffusion models.

\section*{Impact Statement}
This paper presents work whose goal is to advance the field of Machine Learning. It highlights the vulnerability of structured learning pipelines to adversarial examples. It then provides an efficient algorithm to obtain plausible explanations. This has several potential beneficial impacts. Explanations can improve the transparency of these pipelines, build trust in their decisions, and highlight potential biases.

\bibliography{bibliography}
\bibliographystyle{icml2024}


\newpage
\appendix
\onecolumn

\section{Numerical and Theoretical Justifications for the Plausibility Regularization}
\subsection{Numerical Validation of the Approximation of the Encoded Data Distribution by the Prior}
\label{sec:prior_approx}
An important assumption we make in order to derive the plausibility region $\mathcal{D}_\mathcal{Z}$ and the corresponding plausibility regularization $\Omega$ in \cref{sec:plausib_reg} is that the prior over the latent codes $p(\mathbf{z}) = \mathcal{N}\left(0, \mathbf{I}_{n_z}\right)$ is a good approximation of the true encoded data distribution, which can be thought of as the push-forward of $\mathbb{P}_\text{data}$ through the trained encoder $e_\phi$ of the VAE.

This approximation is motivated by the fact that the KL divergence in the training objective of the VAE forces the distribution of the latent codes to match that of the prior $p(\mathbf{z})$. This same approximation is made when using trained VAEs for traditional generative modeling, as the prior is used to sample new data points before passing them through the trained decoder $d_\psi$.

To verify quantitatively if this assumption holds, we compare the percentage of points that should belong to the plausibility region according to the prior (that is, the mass of the region w.r.t $p(\mathbf{z})$) and the percentage of training data points that are actually embedded in the region (that is, the mass of the region w.r.t the empirical encoded data distribution). The table below shows the percentage of plausible points for varying thickness $\kappa$ (in Euclidean distance) of the thickened hypersphere of radius $C_{64}$ when using a VAE with 64-dimensional latent space.

\begin{table*}[htbp]
    \centering
    \caption{Comparison of prior and empirical encoded data distribution}
    \begin{tabular}{lccccccccc}
        \toprule
        & $\kappa=0.0$ & $\kappa=0.25$ & $\kappa=0.5$ & $\kappa=0.75$ & $\kappa=1.0$ & $\kappa=1.25$ & $\kappa=1.5$ & $\kappa=1.75$ & $\kappa=2.0$ \\
        \midrule
        Prior (\%)  & 0.0 & 27.6 & 52.1 & 71.2 & 84.4 & 92.4 & 96.7 & 98.7 & 99.5 \\
        Empirical (\%) & 0.0 & 27.8 & 52.1 & 71.2 & 84.8 & 93.0 & 97.0 & 99.1 & 99.8 \\
        \bottomrule
    \end{tabular}
    \label{tab:my_table}
\end{table*}

Thus, the empirical percentages of training data points embedded in the chosen plausibility region closely match those computed with the statistics of the prior, which further justifies its use as an approximation of the encoded data distribution.

\subsection{Latent Hypersphere as Optimal Plausibility Region}
\label{sec:hyper_proof}

Let $p(\mathbf{z}) = \mathcal{N}\left(0, \mathbf{I}_{n_z}\right)$ be the prior over the latent codes. We use it to approximate the true encoded data distribution, as explained and numerically justified in \cref{sec:prior_approx}. We want to find a plausibility region $\mathcal{D}_\mathcal{Z} \subseteq \mathcal{Z} = \mathbb{R}^{n_z}$ with minimal
\begin{inlinelist}
    \item expected distance to latent codes, and
    \item volume, or Lebesgue measure.
\end{inlinelist} 

These two objectives are conflicting: indeed, the optimal region with respect to the first one only is the whole space, and the optimal region with respect to the second one only is any region that is negligible with respect to the Lebesgue measure. Thus, they cannot be jointly minimized, so we consider the following problem:
\begin{equation*}
        \mathcal{D}_\mathcal{Z} = \argmin_{\mathcal{D} \subseteq \mathcal{Z}} \quad \mathbb{E}_{z\sim p(\mathbf{z})}\left[\ell(z, \mathcal{D})\right] + \eta\,\text{Leb}(\mathcal{D}),
\end{equation*}
where $\eta>0$ is a trade-off coefficient, and $\ell(z, \mathcal{D}) = \min_{z' \in \mathcal{D}}\{\|z-z'\|_2^2\}$ is the squared Euclidean distance of latent vector $z$ to region $\mathcal{D}$.

We assume $\mathcal{D}_\mathcal{Z}$ to be an isotropic region to respect the symmetries of the encoded data distribution (which is approximated by the isotropic centered Gaussian $p(\mathbf{z})$). Thus, one can parameterize it as a union of spheres \text{$\mathcal{D}_\mathcal{Z} = \big\{ z \in \mathcal{Z} \colon \|z\|_2 \in R \big\}$}, where $R\subseteq \mathbb{R}_+$ is the set of radii of the spheres. If we further assume $\mathcal{D}_\mathcal{Z}$ to be connected, which appears logical as $p(\mathbf{z})$ is unimodal, one has that $R$ is an interval of $\mathbb{R}_+$, which we consider to be closed. These assumptions allow us to consider the following simplified optimization objective:
\begin{align}
\label{opt_a_b}
    \argmin_{a,\, b\, \in\, \mathbb{R}_+} \quad &\mathbb{E}_{z\sim p(\mathbf{z})}\left[\ell(z, \mathcal{D}_a^b)\right] + \eta\,\text{Leb}(\mathcal{D}_a^b) \\
    \subto \quad & a\leq b, \notag
\end{align}
where $\mathcal{D}_a^b = \big\{ z \in \mathcal{Z} \colon \|z\|_2 \in \left[a, b\right] \big\}$. Thus, the only candidate regions left are either centered balls (when $a=0$), centered hyperspheres (when $a=b$), and thickened centered hyperspheres (when $0<a<b$).

For $\eta$ large enough in Problem~\eqref{opt_a_b}, the weight of the second term of the optimization objective forces the optimal region to have approximately zero Lebesgue volume. Thus, we can approximate it as $\mathcal{D}_a^a$ (that is, to be a centered, non-thickened hypersphere of radius $a$), and the problem can be approximated as:
\begin{equation}
\label{a_egal_b}
        \argmin_{a\in\mathbb{R}_+} \quad \mathbb{E}_{z\sim p(\mathbf{z})}(a - ||z||_2)^2,
\end{equation}
which is equivalent to:
\begin{equation*}
        \argmin_{a\in\mathbb{R}_+} \quad \mathbb{E}_{X\sim \chi_{n_z}}(a - X)^2,
\end{equation*}
where $\chi_{n_z}$ is a Chi distribution with $n_z$ degrees of freedom since $p(\mathbf{z})$ is a standard $n_z$-dimensional Gaussian. Thus, the solution is the expectation of the $\chi_{n_z}$ distribution, $a= C_{n_z} = \sqrt{2}\frac{\Gamma\left(\frac{n_z+1}{2}\right)}{\Gamma\left(\frac{n_z}{2}\right)} \approx \sqrt{n_z}$, and we find the \textbf{centered hypersphere of radius $C_{n_z}$} as optimal plausibility region.

\paragraph{Numerical valdiation.} We investigate the validity of this approximation numerically. On the one hand, we have:
\begin{align*}
    \text{Leb}(\mathcal{D}_a^b) &= \frac{\pi^{\frac{n_z}{2}}}{\Gamma(n_z/2 + 1)}(b^{n_z}-a^{n_z}),\\
\end{align*}
and on the other hand:
\begin{align*}
    \mathbb{E}_{z\sim p(\mathbf{z})}\left[\ell(z, \mathcal{D}_a^b)\right] &= \int_0^a (a-r)^2(2\pi)^{-\frac{n_z}{2}}e^{-\frac{r^2}{2}}S^{n_z-1}r^{n_z-1}dr + \int_b^{+\infty} (r-b)^2(2\pi)^{-\frac{n_z}{2}}e^{-\frac{r^2}{2}}S^{n_z-1}r^{n_z-1}dr\\
    &\text{(where $S^{n_z-1}= \frac{2\pi^{\frac{n_z}{2}}}{\Gamma\left(\frac{n_z}{2}\right)}$ is the surface of the hypersphere of dimension $n_z-1$ and radius $1$)}\\
    &= a^2\,\mathbb{P}\left(\mathbf{z}\in B(0, a)\right) + b^2\,\mathbb{P}\left(\mathbf{z}\in B(0, b)^c\right) \\
    &- 2a \int_0^a (2\pi)^{-\frac{n_z}{2}}e^{-\frac{r^2}{2}}S^{n_z-1}r^{n_z}dr + \int_0^a (2\pi)^{-\frac{n_z}{2}}e^{-\frac{r^2}{2}}S^{n_z-1}r^{n_z+1}dr \\
    &- 2b\int_b^{+\infty} (2\pi)^{-\frac{n_z}{2}}e^{-\frac{r^2}{2}}S^{n_z-1}r^{n_z}dr + \int_b^{+\infty} (2\pi)^{-\frac{n_z}{2}}e^{-\frac{r^2}{2}}S^{n_z-1}r^{n_z+1}dr\\
    &= a^2\,P\left(\frac{n_z}{2},\, \frac{a^2}{2}\right) + b^2\,Q\left(\frac{n_z}{2},\, \frac{b^2}{2}\right) \\
    &- \frac{2\sqrt{2}}{\Gamma\left(\frac{n_z}{2}\right)} \left[a \left(\Gamma\left(\frac{n_z+1}{2}\right) - \Gamma\left(\frac{n_z+1}{2}, \frac{a^2}{2}\right)\right) + b\, \Gamma\left(\frac{n_z+1}{2}, \frac{b^2}{2}\right) \right]\\
    &+ \frac{2}{\Gamma\left(\frac{n_z}{2}\right)} \left[\Gamma\left(\frac{n_z}{2}+1\right) - \Gamma\left(\frac{n_z}{2}+1, \frac{a^2}{2}\right) + \Gamma\left(\frac{n_z}{2}+1, \frac{b^2}{2}\right) \right],
\end{align*}
where $\Gamma(\cdot\,,\cdot)$ is the upper incomplete gamma function, and $Q(s, x)=\frac{\Gamma(s, x)}{\Gamma(s)}=1-P(s,x)$ with $P(s, x)$ being the cumulative distribution function of a gamma random variable with shape parameter $s$ and scale parameter $1$.

Putting all the pieces together, we have:
\begin{align*}
    \mathbb{E}_{z\sim p(\mathbf{z})}\left[\ell(z, \mathcal{D}_a^b)\right] + \eta\,\text{Leb}(\mathcal{D}_a^b) &= a^2\,P\left(\frac{n_z}{2},\, \frac{a^2}{2}\right) + b^2\,Q\left(\frac{n_z}{2},\, \frac{b^2}{2}\right) \\
    &- 2\sqrt{2}\frac{\Gamma\left(\frac{n_z+1}{2}\right)}{\Gamma\left(\frac{n_z}{2}\right)}\left[ a\,P\left(\frac{n_z+1}{2},\,\frac{a^2}{2}\right) + b\,Q\left(\frac{n_z+1}{2},\,\frac{b^2}{2}\right)\right] \\
    &+ n_z \left[P\left(\frac{n_z}{2}+1,\,\frac{a^2}{2}\right) + Q\left(\frac{n_z}{2}+1,\,\frac{b^2}{2}\right)\right] + \eta\;\frac{\pi^{\frac{n_z}{2}}}{\Gamma(n_z/2 + 1)}(b^{n_z}-a^{n_z}).
\end{align*}

Using the analytical expressions derived above, we verify that our reasoning holds by solving Problem~\eqref{opt_a_b} numerically. To do so, we compute $\Gamma(\cdot)$, $P(\cdot\,,\,\cdot)$ and $Q(\cdot\,,\,\cdot)$ using the functions \texttt{special.gamma}, \texttt{special.gammainc}, and \texttt{special.gammaincc}, respectively, taken from the \texttt{scipy.special} package. We conduct a simple grid search, by taking $10^4$ values for $a$ and $b$ between $0$ and $10$, discarding couples such that $a>b$. For $n_z = 64$, we find that the hypersphere of radius \text{$C_{64}= \sqrt{2}\frac{\Gamma\left(\frac{65}{2}\right)}{\Gamma\left(\frac{64}{2}\right)}\approx 7.97$} is indeed optimal for $\eta \ge 10^{-16}$.

\newpage
\section{Supplementary Material: Numerical Study}
\label{app:numeric}
In this appendix, we provide background details on VAEs as well as the architecture of the VAEs used in our experiments in \cref{app:sec:vae}, and additional experiments on tabular data in \cref{app:sec:tabular}.

\subsection{VAE: Background and Detailed Architecture}
\label{app:sec:vae}
\subsubsection{Background on VAEs}
The encoder $e_\phi$ is chosen to output a Gaussian distribution with a diagonal covariance matrix, that is: 
\begin{equation*}
    e_\phi(\mathbf{z}|x) = \mathcal{N} \left( \mathbf{z};\, \mu_\phi(x),\, \text{Diag}\left[\sigma^2_\phi(x) \right] \right),
\end{equation*}
where $\mu_\phi(x), \; \log\sigma^2_\phi(x)=\log(\sigma^2_\phi(x)) \in \mathbb{R}^{n_z}$ ($\log$ being applied element-wise) are the outputs of the encoder neural network for the input $x \in \mathbb{R}^{n_x}$. We choose the decoder to output an isotropic Gaussian distribution, that is:
\begin{equation*}
    d_\psi(\mathbf{x}|z) = \mathcal{N}(\mathbf{x};\, \mu_\psi(z),\, \mathbf{I}_{n_x})
\end{equation*}
where $\mu_\psi(z) \in \mathbb{R}^{n_x}$ is the output of the decoder neural network for the input $z \in \mathbb{R}^{n_z}$.

Because of our choices for the encoder and decoder, the term inside the expectation in the ELBO \ref{eq:ELBO} is equal to  \text{$-\frac{1}{2}||\mathbf{x}_i -\mu_\psi(z)||_2^2$}, and the KL divergence has a closed-form expression, so that we have (see \citet{odaibo_tutorial_2019} for a guided derivation):

\begin{equation*}
    \mathcal{L}(\phi, \psi, x_i) = -\frac{1}{2}\Exp_{z\sim e_\phi(\mathbf{z}\vert x_i)}\left[\vert \vert x_i - d_\psi(z)\vert\vert _2^2\right] + \frac{1}{2}\sum_{k=1}^{n_z}\left[1 + \log\sigma_\phi^2(x)_k - \sigma_\phi(x)^2_k - \mu_\phi(x)^2_k\right].
\end{equation*}

The expectation in this expression, known as the reconstruction loss (or feature-reconstruction loss to be more precise in our setting), is approximated via a Monte-Carlo estimate during training.

\subsubsection{Architecture for the Warcraft Map Experiments}
All VAEs used in our experiments are implemented in Pytorch.

The encoder's architecture is the following:
\begin{itemize}[topsep=-0.5\parskip, itemsep=-2pt]
    \item Convolutional Layers:
    \begin{itemize}[topsep=-0.5\parskip, itemsep=-2pt]
        \item \texttt{conv1}: A 2D convolutional layer with 32 filters, a kernel size of 4x4, a stride of 2, and padding of 1. It processes the input data, of size $(\textrm{Batch-size}\times3\times96\times96)$ and produces feature maps.
        \item \texttt{conv2}: Another 2D convolutional layer with 64 filters, same kernel size, stride, and padding. It further extracts features.
        \item \texttt{conv3}: A 2D convolutional layer with 128 filters, same parameters as above.
        \item \texttt{conv4}: The final 2D convolutional layer with 256 filters, again with the same settings.
    \end{itemize}
    \item Fully Connected Layers:
    \begin{itemize}[topsep=-0.5\parskip, itemsep=-2pt]
        \item \texttt{fc1}: A fully connected layer with 512 output neurons. It processes the flattened feature maps from the last convolutional layer.
        \item \texttt{fc21} and \texttt{fc22}: These fully connected layers produce two vectors: $\mu_\phi(x)$ and $\log\sigma^2_\phi(x)$ (representing the log variance). These vectors are used to sample points in the latent space.
    \end{itemize}
    \item Activation Functions: we use ReLU activation functions after each layer except after \texttt{fc21} and \texttt{fc22}.
\end{itemize}

The decoder's architecture is the following:
\begin{itemize}[topsep=-0.5\parskip, itemsep=-2pt]
    \item Fully Connected Layers:
    \begin{itemize}
        \item \texttt{fc10}: A fully connected layer with 512 output neurons, that takes the latent vector as input and produces a feature representation.
        \item \texttt{fc20}: Another fully connected layer that expands the feature representation back to the original flattened shape (6x6x256).
    \end{itemize}
    \item Transposed Convolutional Layers (also known as deconvolutional layers):
    \begin{itemize}[topsep=-0.5\parskip, itemsep=-2pt]
        \item \texttt{deconv1}: A transposed 2D convolutional layer with 128 filters, kernel size 4x4, stride 2, and padding 1. It upsamples the feature maps.
        \item \texttt{deconv2}: Another transposed 2D convolutional layer with 64 filters, same parameters.
        \item \texttt{deconv3}: A transposed 2D convolutional layer with 32 filters, same settings.
        \item \texttt{deconv4}: The final transposed 2D convolutional layer with 3 filters (giving RGB channels), same parameters. It generates the reconstructed data.
    \end{itemize}
    \item Activation Functions: we use ReLU activation functions after each layer except after \texttt{deconv4}, where we use a Sigmoid function.
\end{itemize}

\subsection{Repeated Experiments on Tabular Data}
\label{app:sec:tabular}

\paragraph{Experimental setting.} We perform repeated experiments generating counterfactual explanations of decisions and measure the sensitivity of our method to changes in the problem settings. We measure the total number of iterations of \texttt{CF-OPT} when varying the number of features, the number of decision variables, and the complexity of the prediction model. 
\begin{remark}
    We do not measure the sensitivity of our method to the number of training points in the dataset since the computational complexity of calculating an explanation with \model{CF-OPT} does not increase with the size of the training dataset. This is a clear improvement compared to \citet{Forel2023}.
\end{remark}

\subsubsection{Shortest paths on a grid.}

We apply our explanation method to the shortest-path problem on a square grid introduced by \citet{Elmachtoub2022} and used in several other works \citep{Elmachtoub2020, tang2022pyepo}. The structured learning pipeline consists of a contextual vector $x\in \mathbb{R}^{n_x}$, a linear regression model $\varphi$ predicting costs \text{$\varphi(x)=\theta \in \mathbb{R}^{n_y}$} on the edges of a \text{$(N\times N)$} grid, so we have \text{$n_y = 2N(N-1)$}, and a linear optimization model computing the shortest path from the upper left node to the lower right node of the grid. The optimization problem is still cast as a linear program and solved with Gurobi.

The data is generated following \citet{tang2022pyepo}. The cost of an edge of the graph representing the square grid is given by
\begin{equation}
    \label{eq:pyepo_cost}
    (\theta_i)_j = \left[ \frac{1}{3.5^4}\left( \frac{1}{\sqrt{n_x}}\left( B x_i \right)_j + 3\right)^4 +1 \right]\cdot \epsilon_{ij},
\end{equation}
where \text{$B\in \mathbb{R}^{n_y \times n_x}$} is a random matrix whose each element follows a Bernoulli distribution with parameter $0.5$, and \text{$\epsilon_{ij} \sim U(0.5, 1.5)$}. The contextual vectors $x_i$ follow a standard multivariate Gaussian $\mathcal{N}(\mathbf{0}, \mathbf{I}_{n_x})$.

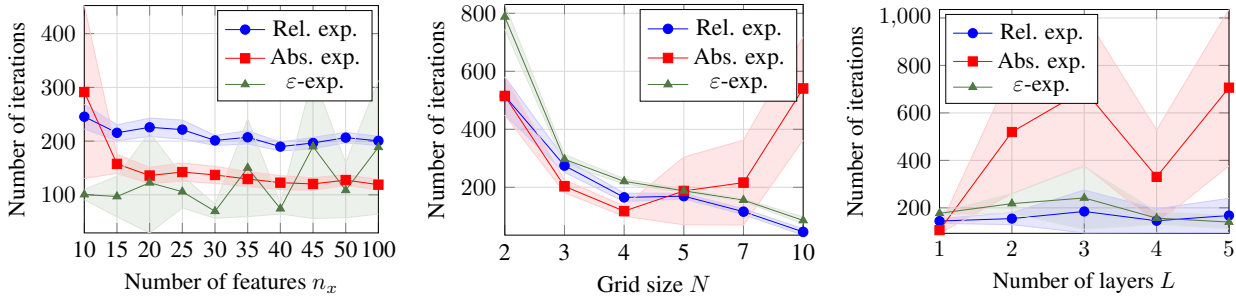
\begin{figure}[ht]
    \vspace{-1mm}
    \centering
    \begin{subfigure}[t]{0.32\linewidth}
        \resizebox{0.99\linewidth}{!}{\begin{tikzpicture}
    \begin{axis}[
            height = 5cm,
            width  = 6cm,
            enlarge x limits = 0,
            enlarge y limits = 0,
            xlabel = {Number of features $n_x$},
            ylabel = {Number of iterations},
            symbolic x coords ={10, 15, 20, 25, 30, 35, 40, 45, 50, 100},
            xtick = data
        ]
        \addplot+[relStyle] table [x index = {0}, y index = {1}, col sep=comma]{plots/csv/spg/varying_context_dim.csv};
        \addplot+[name path=A, blue!20, mark=none, forget plot] table [x index = {0}, y index = {2}, col sep=comma]{plots/csv/spg/varying_context_dim.csv};
        \addplot+[name path=B, blue!20, mark=none, forget plot] table [x index = {0}, y index = {3}, col sep=comma]{plots/csv/spg/varying_context_dim.csv};
        \addplot[blue, fill opacity=0.1] fill between[of=A and B];

        
        \addplot+[absStyle]  table [x index = {0}, y index = {4}, col sep=comma]{plots/csv/spg/varying_context_dim.csv};
        \addplot+[name path=A, red!20, mark=none, forget plot] table [x index = {0}, y index = {5}, col sep=comma]{plots/csv/spg/varying_context_dim.csv};
        \addplot+[name path=B, red!20, mark=none, forget plot] table [x index = {0}, y index = {6}, col sep=comma]{plots/csv/spg/varying_context_dim.csv};
        \addplot[red, fill opacity=0.1, forget plot] fill between[of=A and B, forget plot];

        \addplot+[epsiStyle] table [x index = {0}, y index = {7}, col sep=comma]{plots/csv/spg/varying_context_dim.csv};
        \addplot+[name path=A, darkgreen!20, mark=none, forget plot] table [x index = {0}, y index = {8}, col sep=comma]{plots/csv/spg/varying_context_dim.csv};
        \addplot+[name path=B, darkgreen!20, mark=none, forget plot] table [x index = {0}, y index = {9}, col sep=comma]{plots/csv/spg/varying_context_dim.csv};
        \addplot[darkgreen, fill opacity=0.1, forget plot] fill between[of=A and B];
        \legend{Rel. exp., , Abs. exp., $\varepsilon$-exp.}
    \end{axis}
\end{tikzpicture}}
        \caption{Number of iterations for varying contextual dimension}
        \label{fig:comp_context}
    \end{subfigure}
    \begin{subfigure}[t]{0.32\linewidth}
        \resizebox{0.99\linewidth}{!}{\begin{tikzpicture}
    \begin{axis}[
            height = 5cm,
            width  = 6cm,
            enlarge x limits = 0,
            enlarge y limits = 0,
            xlabel = {Grid size $N$},
            ylabel = {Number of iterations},
            symbolic x coords ={2, 3, 4, 5, 7, 10},
            xtick = data
        ]

        \addplot+[relStyle] table [x index = {0}, y index = {1}, col sep=comma]{plots/csv/spg/varying_grid_size.csv};
        \addlegendentry{Rel. exp.}
        \addplot+[name path=A, blue!20, mark=none, forget plot] table [x index = {0}, y index = {2}, col sep=comma]{plots/csv/spg/varying_grid_size.csv};
        \addplot+[name path=B, blue!20, mark=none, forget plot] table [x index = {0}, y index = {3}, col sep=comma]{plots/csv/spg/varying_grid_size.csv};
        \addplot[blue, fill opacity=0.1, forget plot] fill between[of=A and B];

        \addplot+[absStyle]  table [x index = {0}, y index = {4}, col sep=comma]{plots/csv/spg/varying_grid_size.csv};
        \addlegendentry{Abs. exp.}
        \addplot+[name path=A, red!20, mark=none, forget plot] table [x index = {0}, y index = {5}, col sep=comma]{plots/csv/spg/varying_grid_size.csv};
        \addplot+[name path=B, red!20, mark=none, forget plot] table [x index = {0}, y index = {6}, col sep=comma]{plots/csv/spg/varying_grid_size.csv};
        \addplot[red, fill opacity=0.1, forget plot] fill between[of=A and B];

        \addplot+[epsiStyle] table [x index = {0}, y index = {7}, col sep=comma]{plots/csv/spg/varying_grid_size.csv};
        \addlegendentry{$\varepsilon$-exp.}
        \addplot+[name path=A, darkgreen!20, mark=none, forget plot] table [x index = {0}, y index = {8}, col sep=comma]{plots/csv/spg/varying_grid_size.csv};
        \addplot+[name path=B, darkgreen!20, mark=none, forget plot] table [x index = {0}, y index = {9}, col sep=comma]{plots/csv/spg/varying_grid_size.csv};
        \addplot[darkgreen, fill opacity=0.1, forget plot] fill between[of=A and B];

    \end{axis}
\end{tikzpicture}}
        \caption{Number of iterations for varying decision dimension}
        \label{fig:comp_grid}
    \end{subfigure}
    \begin{subfigure}[t]{0.32\linewidth}
        \resizebox{0.99\linewidth}{!}{\begin{tikzpicture}
    \begin{axis}[
            height = 5cm,
            width  = 6cm,
            enlarge x limits = 0,
            enlarge y limits = 0,
            xlabel = {Number of layers $L$},
            ylabel = {Number of iterations},
            legend style={at={(0.02,0.98)},anchor=north west}
        ]

        \addplot+[relStyle] table [x index = {0}, y index = {1}, col sep=comma]{plots/csv/spg/varying_depth.csv};
        \addlegendentry{Rel. exp.}
        \addplot+[name path=A, blue!20, mark=none, forget plot] table [x index = {0}, y index = {2}, col sep=comma]{plots/csv/spg/varying_depth.csv};
        \addplot+[name path=B, blue!20, mark=none, forget plot] table [x index = {0}, y index = {3}, col sep=comma]{plots/csv/spg/varying_depth.csv};
        \addplot[blue, fill opacity=0.1, forget plot] fill between[of=A and B];

        \addplot+[absStyle]  table [x index = {0}, y index = {4}, col sep=comma]{plots/csv/spg/varying_depth.csv};
        \addlegendentry{Abs. exp.}
        \addplot+[name path=A, red!20, mark=none, forget plot] table [x index = {0}, y index = {5}, col sep=comma]{plots/csv/spg/varying_depth.csv};
        \addplot+[name path=B, red!20, mark=none, forget plot] table [x index = {0}, y index = {6}, col sep=comma]{plots/csv/spg/varying_depth.csv};
        \addplot[red, fill opacity=0.1, forget plot] fill between[of=A and B];

        \addplot+[epsiStyle] table [x index = {0}, y index = {7}, col sep=comma]{plots/csv/spg/varying_depth.csv};
        \addlegendentry{$\varepsilon$-exp.}
        \addplot+[name path=A, darkgreen!20, mark=none, forget plot] table [x index = {0}, y index = {8}, col sep=comma]{plots/csv/spg/varying_depth.csv};
        \addplot+[name path=B, darkgreen!20, mark=none, forget plot] table [x index = {0}, y index = {9}, col sep=comma]{plots/csv/spg/varying_depth.csv};
        \addplot[darkgreen, fill opacity=0.1, forget plot] fill between[of=A and B];

    \end{axis}
\end{tikzpicture}}
        \caption{Number of iterations for varying model complexity (depth of feed-forward neural network predictor)}
        \label{fig:comp_depth}
    \end{subfigure}
    \vspace{-1mm}
    \caption{Experiment results with tabular data on the shortest paths on a grid problem, showing the sensitivity of our algorithm to the number of (a)~features, (b)~decisions, and (c)~layers}
\end{figure}

\paragraph{Sensitivity to contextual dimension.} First, we analyze the effect of the number of contextual features on the computation time for each explanation type. For each number of contextual features $n_x \in \{5, 10, 25, 50, 100, 500\}$, we generate data: contextual vectors \text{$(x_i)_{i=1}^{2000}$}, associated costs \text{$(\theta_i)_{i=1}^{2000}$} and shortest paths \text{$(y_i)_{i=1}^{2000}$}. The grid size if fixed to $(5\times 5)$ (that is, $N=5$ and $n_y = 40$). Then, we train a linear regression model with the SPO+ loss \citep{Elmachtoub2022} for 70 epochs (until convergence) on the $1000$ first data (train set), with Adam \citep{kingma_adam_2017} and a $3\times 10^{-4}$ learning rate.

In each setting, we perform $100$ explanations of each type, on $100$ random pairs $(x_i, y_j)$, belonging to the $1000$ last generated data (test set). $x_i$ plays the role of $x_0$ (initial context), and $y_j$ plays the role of $\yAlt$ (alternative solution). We check that $y_i \neq y_j$ for relative and absolute explanations to ensure that the explanation problem is not trivial (\text{$\varepsilon$-explanations} do not require an $\yAlt$ anyway). \text{$\varepsilon$-explanations} are performed with $\varepsilon=1$. We use a step size $\gamma=0.1$, maximum number of iterations $K=6000$, maximum number of non-improving iterations $c_{\max} = 10$, and update tolerance $u=0.9$. The results are displayed in \cref{fig:comp_context}, and show that explanations can be obtained efficiently even for large numbers of features.

\paragraph{Sensitivity to decision dimension.} We now analyze the effect of the size of the costs grid (and thus the number of decision features $n_y$) on the computational time for each explanation type. The data generation and training processes are unchanged w.r.t. the above experiment. We choose a fixed number of contextual features $n_x = 10$, and experiment for \text{$N \in \{2, 3, 4, 5, 7, 10\}$} (corresponding to \text{$n_y \in \{ 4, 12, 24, 40, 84, 180\}$}). The results are shown in \cref{fig:comp_grid}. The figure shows that obtaining relative and $\varepsilon$-explanations scales well in the number of decisions, contrary to absolute explanations. This is caused by the fact that the grid size is the source of the combinatorial explosion of the number of solutions.

\paragraph{Sensitivity to prediction model's complexity.} Then, we analyze the effect of the prediction model's complexity, by using a fully connected feed-forward ReLU neural network as $\varphi_L$, and observing the evolution of computation times for different numbers of layers (denoted $L$). The number of contextual features is fixed to $n_x = 10$, and the grid size to $(5\times 5)$ (that is, $N=5$ and $n_y=40$). The first layer has input dimension $n_x$ and output dimension $n_x$ ($n_y$ if \text{$L=1$}, which boils down to the linear regression model), and all other layers have input and output dimension equal to $n_x$ except the output layer, which has output dimension $n_y$. The train/test dataset is generated only once, with $10000$ data points in the first and $1000$ in the latter. For \text{$L\in \{ 1, 2, 3, 4, 5 \}$}, we train $\varphi_L$, whose architecture is described above, until convergence (before overfitting) on the train set with Adam and a $3\times10^{-4}$ learning rate. The results are shown in \cref{fig:comp_depth}, and the number of iterations needed to obtain explanations is almost independent of the model complexity for relative and $\varepsilon$ explanations.

\begin{minipage}{0.5\textwidth}
    \paragraph{Varying $\varepsilon$ for \text{$\varepsilon$-explanations.}} We finally analyze the effect of $\varepsilon$ on \text{$\varepsilon$-explanation} tasks by observing the distance of the obtained explanation to the initial context $||\xAlt - x_0||_2^2$. We use a single dataset (train and test), generated with $n_x = 10$ contextual features, a $(5\times 5)$ grid (that is, $N=5$ and $n_y=40$), and the same generating routine as before. The linear regression model is trained with the same process as described above. For $\varepsilon \in \{0.2, 0.5, 0.7, 1, 1.5, 2, 3\}$, we perform $100$ \text{$\varepsilon$-explanations} on $100$ random initial contexts $x_0$, taken from the test dataset. The results are displayed in \cref{fig:plot_epsilon} and show, as expected, that \text{$\varepsilon$-explanations} move away from the initial context as $\varepsilon$ increases.
\end{minipage}\hfill%
\begin{minipage}{0.4\textwidth}
    \resizebox{\linewidth}{!}{
    \begin{tikzpicture}

    \pgfplotsset{
        scale only axis,
        xmin=0.2, xmax=3
    }
    \begin{axis}[
            height = 4cm,
            width  = 8cm,
            enlarge x limits = 0,
            enlarge y limits = 0,
            ymin=0, ymax=10,
            xlabel = {$\varepsilon$},
            ylabel = {Squared $\ell^2$ distance to initial context},
            symbolic x coords ={0.2, 0.5, 0.7, 1, 1.5, 2, 3},
            xtick = data,
            legend pos=north west
        ]

        \addplot+[epsproxStyle] table [x index = {0}, y index = {1}, col sep=comma]{plots/csv/spg/varying_epsilon.csv};
        \addlegendentry{$||\xAlt_\varepsilon -  x_0||^2_2$ (mean)}
        \addplot+[name path=A, brown!20, mark=none, forget plot] table [x index = {0}, y index = {2}, col sep=comma]{plots/csv/spg/varying_epsilon.csv};
        \addplot+[name path=B, brown!20, mark=none, forget plot] table [x index = {0}, y index = {3}, col sep=comma]{plots/csv/spg/varying_epsilon.csv};
        \addplot[brown, fill opacity=0.1, forget plot] fill between[of=A and B];

    \end{axis}
\end{tikzpicture}}
    
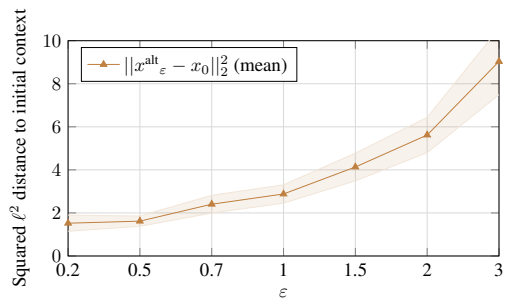
\captionof{figure}{Proximity of $\varepsilon$-explanation $\xAlt_\varepsilon$ to initial context $x_0$ for varying $\varepsilon$, measured with squared $\ell^2$.}
    \label{fig:plot_epsilon}
\end{minipage}

\subsubsection{Contextual Multi-Dimensional Knapsack}
We evaluate our approach on the contextual multi-dimensional knapsack problem presented in \citet{tang2022pyepo}, in which the decision-maker maximizes the expected rewards of items to select subject to weight constraints. The relationship between the features and rewards is also given by \cref{eq:pyepo_cost}. The structured learning pipeline is identical to the one in the previous experiment. The optimization model is formulated as a binary model, in which each variable indicates whether an item is selected or not, cast as an integer linear program and solved using Gurobi.

\begin{figure}[ht]
    \vspace{-1mm}
    \centering
    \begin{subfigure}[t]{0.32\linewidth}
        \resizebox{0.99\linewidth}{!}{\begin{tikzpicture}
    \begin{axis}[
            height = 5cm,
            width  = 6cm,
            enlarge x limits = 0,
            enlarge y limits = 0,
            xlabel = {Number of features $n_x$},
            ylabel = {Number of iterations},
            symbolic x coords ={2, 3, 5, 7, 10, 12, 15, 20, 25, 30},
            xtick = data
        ]
        \addplot+[relStyleKnap] table [x index = {0}, y index = {1}, col sep=comma]{plots/csv/knapsack/result_feat_new.csv};
        \addplot+[name path=A, MidnightBlue!20, mark=none, forget plot] table [x index = {0}, y index = {2}, col sep=comma]{plots/csv/knapsack/result_feat_new.csv};
        \addplot+[name path=B, MidnightBlue!20, mark=none, forget plot] table [x index = {0}, y index = {3}, col sep=comma]{plots/csv/knapsack/result_feat_new.csv};
        \addplot[MidnightBlue, fill opacity=0.1] fill between[of=A and B];

        
        \addplot+[absStyleKnap]  table [x index = {0}, y index = {4}, col sep=comma]{plots/csv/knapsack/result_feat_new.csv};
        \addplot+[name path=A, BrickRed!20, mark=none, forget plot] table [x index = {0}, y index = {5}, col sep=comma]{plots/csv/knapsack/result_feat_new.csv};
        \addplot+[name path=B, BrickRed!20, mark=none, forget plot] table [x index = {0}, y index = {6}, col sep=comma]{plots/csv/knapsack/result_feat_new.csv};
        \addplot[BrickRed, fill opacity=0.1, forget plot] fill between[of=A and B, forget plot];

        \addplot+[epsiStyleKnap] table [x index = {0}, y index = {7}, col sep=comma]{plots/csv/knapsack/result_feat_new.csv};
        \addplot+[name path=A, OliveGreen!20, mark=none, forget plot] table [x index = {0}, y index = {8}, col sep=comma]{plots/csv/knapsack/result_feat_new.csv};
        \addplot+[name path=B, OliveGreen!20, mark=none, forget plot] table [x index = {0}, y index = {9}, col sep=comma]{plots/csv/knapsack/result_feat_new.csv};
        \addplot[OliveGreen, fill opacity=0.1, forget plot] fill between[of=A and B];
        \legend{Rel. exp., , Abs. exp., $\varepsilon$-exp.}
    \end{axis}
\end{tikzpicture}}
        \caption{Number of iterations for varying contextual dimension}
        \label{fig:knap_feat}
    \end{subfigure}
    \begin{subfigure}[t]{0.32\linewidth}
        \resizebox{0.99\linewidth}{!}{\begin{tikzpicture}
    \begin{axis}[
            height = 5cm,
            width  = 6cm,
            enlarge x limits = 0,
            enlarge y limits = 0,
            xlabel = {Number of items $m$},
            ylabel = {Number of iterations},
            symbolic x coords ={5, 6, 7, 8, 9, 10, 15, 20, 25, 30},
            xtick = data,
            legend pos=north west
        ]
        \addplot+[relStyleKnap] table [x index = {0}, y index = {1}, col sep=comma]{plots/csv/knapsack/result_item_new.csv};
        \addplot+[name path=A, MidnightBlue!20, mark=none, forget plot] table [x index = {0}, y index = {2}, col sep=comma]{plots/csv/knapsack/result_item_new.csv};
        \addplot+[name path=B, MidnightBlue!20, mark=none, forget plot] table [x index = {0}, y index = {3}, col sep=comma]{plots/csv/knapsack/result_item_new.csv};
        \addplot[MidnightBlue, fill opacity=0.1] fill between[of=A and B];

        \addplot+[absStyleKnap]  table [x index = {0}, y index = {4}, col sep=comma]{plots/csv/knapsack/result_item_new.csv};
        \addplot+[name path=A, BrickRed!20, mark=none, forget plot] table [x index = {0}, y index = {5}, col sep=comma]{plots/csv/knapsack/result_item_new.csv};
        \addplot+[name path=B, BrickRed!20, mark=none, forget plot] table [x index = {0}, y index = {6}, col sep=comma]{plots/csv/knapsack/result_item_new.csv};
        \addplot[BrickRed, fill opacity=0.1, forget plot] fill between[of=A and B, forget plot];

        \addplot+[epsiStyleKnap] table [x index = {0}, y index = {7}, col sep=comma]{plots/csv/knapsack/result_item_new.csv};
        \addplot+[name path=A, OliveGreen!20, mark=none, forget plot] table [x index = {0}, y index = {8}, col sep=comma]{plots/csv/knapsack/result_item_new.csv};
        \addplot+[name path=B, OliveGreen!20, mark=none, forget plot] table [x index = {0}, y index = {9}, col sep=comma]{plots/csv/knapsack/result_item_new.csv};
        \addplot[OliveGreen, fill opacity=0.1, forget plot] fill between[of=A and B];
        \legend{Rel. exp., , Abs. exp., $\varepsilon$-exp.}
    \end{axis}
\end{tikzpicture}}
        \caption{Number of iterations for varying number of items}
        \label{fig:knap_item}
    \end{subfigure}
    \begin{subfigure}[t]{0.32\linewidth}
        \resizebox{0.99\linewidth}{!}{\begin{tikzpicture}
    \begin{axis}[
            height = 5cm,
            width  = 6cm,
            enlarge x limits = 0,
            enlarge y limits = 0,
            xlabel = {Number of layers $L$},
            ylabel = {Number of iterations},
            symbolic x coords ={1, 2, 3, 4, 5},
            xtick = data,
            legend pos=south east
        ]
        \addplot+[relStyleKnap] table [x index = {0}, y index = {1}, col sep=comma]{plots/csv/knapsack/result_depth_new.csv};
        \addplot+[name path=A, MidnightBlue!20, mark=none, forget plot] table [x index = {0}, y index = {2}, col sep=comma]{plots/csv/knapsack/result_depth_new.csv};
        \addplot+[name path=B, MidnightBlue!20, mark=none, forget plot] table [x index = {0}, y index = {3}, col sep=comma]{plots/csv/knapsack/result_depth_new.csv};
        \addplot[MidnightBlue, fill opacity=0.1] fill between[of=A and B];

        
        \addplot+[absStyleKnap]  table [x index = {0}, y index = {4}, col sep=comma]{plots/csv/knapsack/result_depth_new.csv};
        \addplot+[name path=A, BrickRed!20, mark=none, forget plot] table [x index = {0}, y index = {5}, col sep=comma]{plots/csv/knapsack/result_depth_new.csv};
        \addplot+[name path=B, BrickRed!20, mark=none, forget plot] table [x index = {0}, y index = {6}, col sep=comma]{plots/csv/knapsack/result_depth_new.csv};
        \addplot[BrickRed, fill opacity=0.1, forget plot] fill between[of=A and B, forget plot];

        \addplot+[epsiStyleKnap] table [x index = {0}, y index = {7}, col sep=comma]{plots/csv/knapsack/result_depth_new.csv};
        \addplot+[name path=A, OliveGreen!20, mark=none, forget plot] table [x index = {0}, y index = {8}, col sep=comma]{plots/csv/knapsack/result_depth_new.csv};
        \addplot+[name path=B, OliveGreen!20, mark=none, forget plot] table [x index = {0}, y index = {9}, col sep=comma]{plots/csv/knapsack/result_depth_new.csv};
        \addplot[OliveGreen, fill opacity=0.1, forget plot] fill between[of=A and B];
        \legend{Rel. exp., , Abs. exp., $\varepsilon$-exp.}
    \end{axis}
\end{tikzpicture}}
        \caption{Number of iterations for varying model complexity (depth of feed-forward neural network predictor)}
        \label{fig:knap_depth}
    \end{subfigure}
    \vspace{-1mm}
    \caption{Experiment results with tabular data on the contextual multi-dimensional knapsack problem, showing the sensitivity of our algorithm to the number of (a)~features, (b)~items, and (c)~layers}
\end{figure}
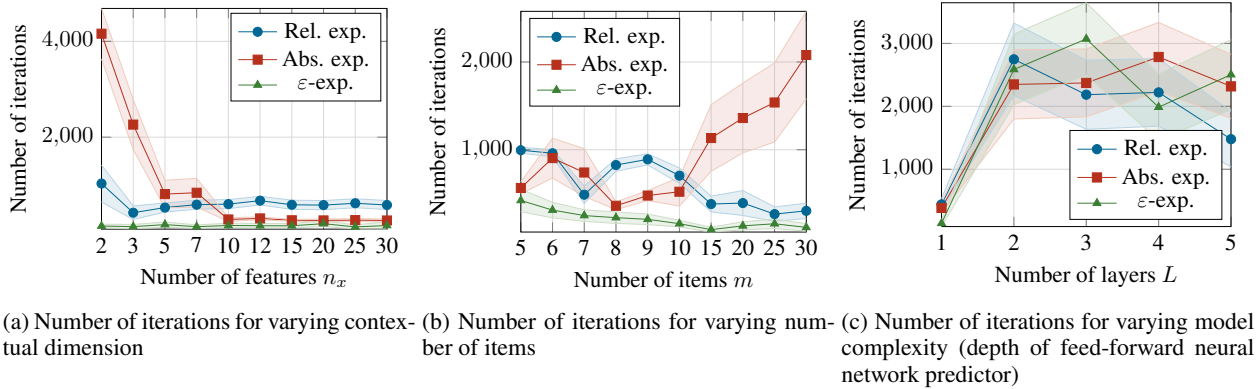

\paragraph{Sensitivity to contextual dimension.} First, we analyze the effect of the number of contextual features on the computation time for each explanation type on the knapsack problem. For each number of contextual features \text{$n_x \in \{2, 3, 5, 7, 10, 12, 15, 20, 25, 30\}$}, we generate data: contextual vectors \text{$(x_i)_{i=1}^{6000}$}, associated rewards \text{$(\theta_i)_{i=1}^{6000}$} and optimal solutions \text{$(y_i)_{i=1}^{6000}$}. The number of items is fixed to $m=16$. Then, we train a linear regression model with the SPO+ loss \citep{Elmachtoub2022} for 30 epochs (until convergence) on the $5000$ first data (train set), with Adam \citep{kingma_adam_2017} and a $1\times 10^{-3}$ learning rate.

In each setting, we perform $100$ explanations of each type, on $100$ random pairs $(x_i, y_j)$, belonging to the $1000$ last generated data (test set). $x_i$ plays the role of $x_0$ (initial context), and $y_j$ plays the role of $\yAlt$ (alternative solution). We check that $y_i \neq y_j$ for relative and absolute explanations to ensure that the explanation problem is not trivial (\text{$\varepsilon$-explanations} do not require an $\yAlt$ anyway). \text{$\varepsilon$-explanations} are performed with $\varepsilon=0.1$. We use a step size $\gamma=0.01$, maximum number of iterations $K=6000$, maximum number of non-improving iterations $c_{\max} = 10$, and corresponding update tolerance $u=0.9$. The results are displayed in \cref{fig:knap_feat}. They show once again that explanations can be obtained efficiently even for a larger number of features. The absolute explanations, however, require more iterations on average when the number of features is very low: this is caused by the fact that the prediction model being linear, the set of rewards it can span is limited by its rank (and thus by its input dimension), leading to infeasible absolute explanation tasks which increase the average computation time by taking $K=6000$ iterations (the maximum) to stop.

\paragraph{Sensitivity to the number of items.} We now analyze the effect of the number of items $m$ on the computational time for each explanation type. The data generation and training processes are unchanged w.r.t. the above experiment. We choose a fixed number of contextual features $n_x = 5$ and experiment for \text{$m \in \{5, 6, 7, 8, 9, 10, 15, 20, 25, 30\}$}. The results are shown in \cref{fig:knap_item}. The figure highlights that obtaining relative and $\varepsilon$-explanations scales well in the number of items, contrary to absolute explanations This is caused by the fact that the number of items greatly improves the combinatorial difficulty of the problem.

\paragraph{Sensitivity to the prediction model's complexity.} Then, we analyze the effect of the prediction model's complexity by using a fully connected feed-forward ReLU neural network as $\varphi_L$, and observing the evolution of computation times for different numbers of layers (denoted $L$). The architecture is similar to the one used in the shortest paths on a grid experiment. The train/test dataset is generated only once, with $5000$ data points in the first and $1000$ in the latter. For \text{$L\in \{ 1, 2, 3, 4, 5 \}$}, we train $\varphi_L$, whose architecture is described above, until convergence (before overfitting) on the train set with Adam and a $1\times10^{-3}$ learning rate. 

The results are shown in \cref{fig:knap_depth}, which shows that the number of iterations needed to obtain explanations for the knapsack problem is almost independent of the model complexity when the number of layers exceeds one. When only one layer is used (the predictor then boils down to a linear regression), the average number of iterations required is significantly lower for all types of explanations.

\begin{minipage}{0.5\textwidth}
    \paragraph{Varying $\varepsilon$ for \text{$\varepsilon$-explanations.}} We finally analyze the effect of $\varepsilon$ on \text{$\varepsilon$-explanation} tasks for the knapsack problem by observing the distance of the obtained explanation to the initial context $||\xAlt - x_0||_2^2$. We use a single dataset (train and test) generated with $n_x = 5$ contextual features, $m=16$ items, and the same generating routine as before. The linear regression model is trained using the same process as described above. For $\varepsilon \in \{0.2, 0.5, 0.7, 1, 1.5, 2, 3\}$, we perform $100$ \text{$\varepsilon$-explanations} on $100$ random initial contexts $x_0$, taken from the test dataset. The results are displayed in \cref{fig:plot_knapsack_epsilon} and show, as expected and as for the shortest paths on a grid experiment, that \text{$\varepsilon$-explanations} move away from the initial context as $\varepsilon$ increases.
\end{minipage}\hfill%
\begin{minipage}{0.4\textwidth}
    \resizebox{\linewidth}{!}{
    \begin{tikzpicture}

    \pgfplotsset{
        scale only axis,
        xmin=0.2, xmax=3
    }
    \begin{axis}[
            height = 4cm,
            width  = 8cm,
            enlarge x limits = 0,
            enlarge y limits = 0,
            ymin=0,
            xlabel = {$\varepsilon$},
            ylabel = {Squared $\ell^2$ distance to initial context},
            symbolic x coords ={0.1, 0.2, 0.5, 0.7, 1, 1.5, 2, 3},
            xtick = data,
            legend pos=north west
        ]

        \addplot+[epsproxStyle] table [x index = {0}, y index = {1}, col sep=comma]{plots/csv/knapsack/result_eps_new.csv};
        \addlegendentry{$||\xAlt_\varepsilon -  x_0||^2_2$ (mean)}
        \addplot+[name path=A, brown!20, mark=none, forget plot] table [x index = {0}, y index = {2}, col sep=comma]{plots/csv/knapsack/result_eps_new.csv};
        \addplot+[name path=B, brown!20, mark=none, forget plot] table [x index = {0}, y index = {3}, col sep=comma]{plots/csv/knapsack/result_eps_new.csv};
        \addplot[brown, fill opacity=0.1, forget plot] fill between[of=A and B];

    \end{axis}
\end{tikzpicture}}
    
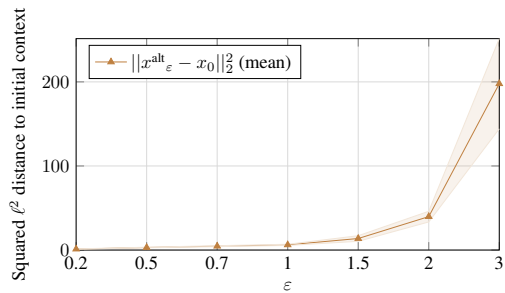
\captionof{figure}{Proximity of $\varepsilon$-explanation $\xAlt_\varepsilon$ to initial context $x_0$ for varying $\varepsilon$, measured with squared $\ell^2$.}
    \label{fig:plot_knapsack_epsilon}
\end{minipage}

\newpage
\section{Examples of Explanations for the Shortest Paths on Warcraft Maps Pipeline}
\label{app:sec:examples}

\begin{figure}
    \centering
    \begin{subfigure}[b]{0.22\textwidth}
        \includegraphics[width=\textwidth]{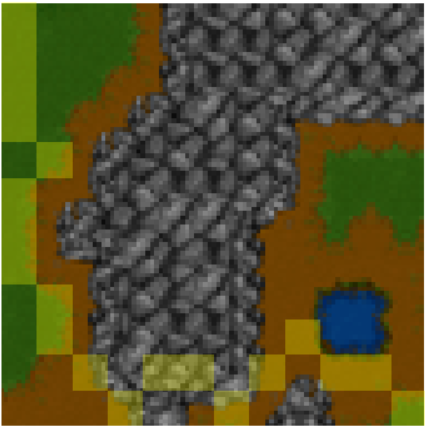}

    \end{subfigure}
    \hfill
    \begin{subfigure}[b]{0.22\textwidth}
        \includegraphics[width=\textwidth]{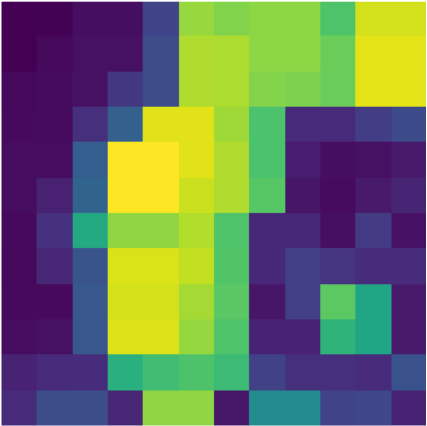}
    \end{subfigure}
    \hfill
    \begin{subfigure}[b]{0.22\textwidth}
        \includegraphics[width=\textwidth]{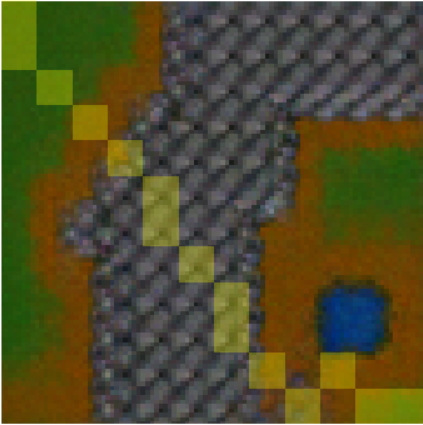}
    \end{subfigure}
    \hfill
    \begin{subfigure}[b]{0.22\textwidth}
        \includegraphics[width=\textwidth]{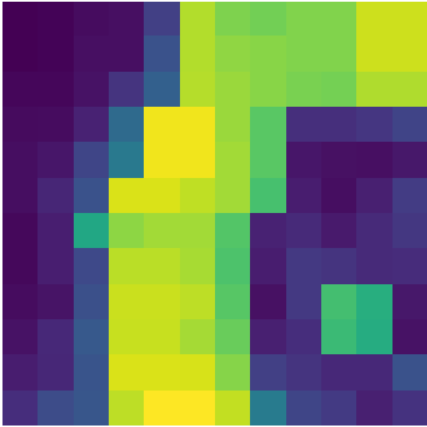}
    \end{subfigure}
    \\
    \begin{subfigure}[b]{0.22\textwidth}
        \includegraphics[width=\textwidth]{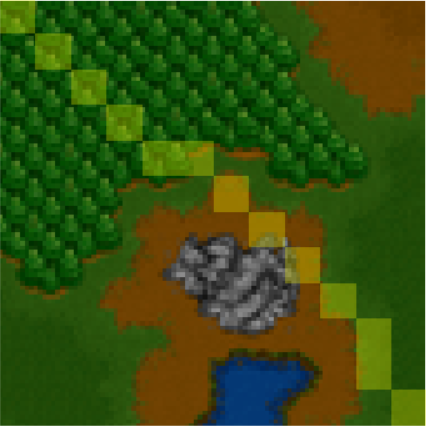}
    \end{subfigure}
    \hfill
    \begin{subfigure}[b]{0.22\textwidth}
        \includegraphics[width=\textwidth]{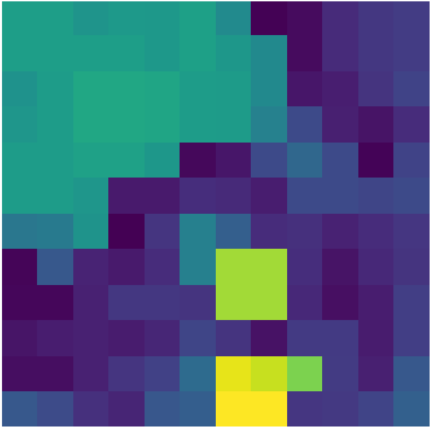}
    \end{subfigure}
    \hfill
    \begin{subfigure}[b]{0.22\textwidth}
        \includegraphics[width=\textwidth]{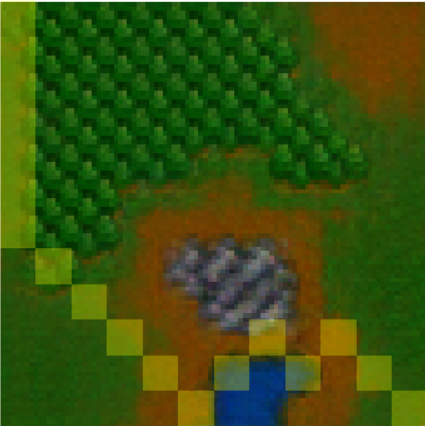}
    \end{subfigure} 
    \hfill
    \begin{subfigure}[b]{0.22\textwidth}
        \includegraphics[width=\textwidth]{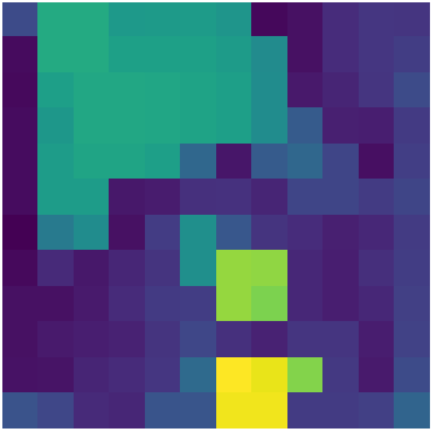}
    \end{subfigure}
    \\
    \begin{subfigure}[b]{0.22\textwidth}
        \includegraphics[width=\textwidth]{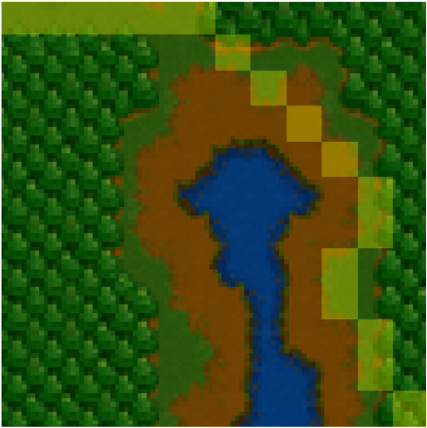}
    \end{subfigure}
    \hfill
    \begin{subfigure}[b]{0.22\textwidth}
        \includegraphics[width=\textwidth]{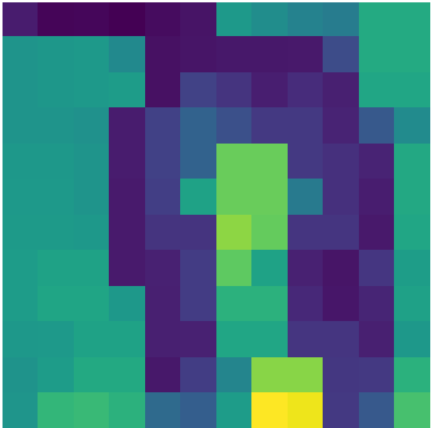}
    \end{subfigure}
    \hfill
    \begin{subfigure}[b]{0.22\textwidth}
        \includegraphics[width=\textwidth]{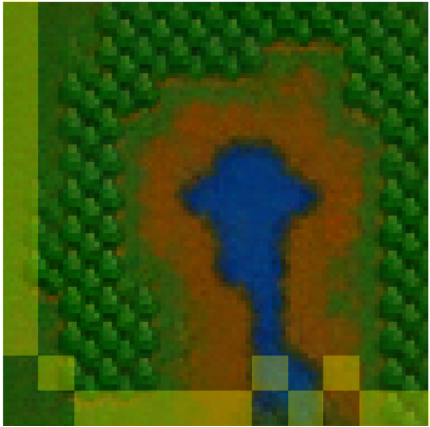}
    \end{subfigure}
    \hfill
    \begin{subfigure}[b]{0.22\textwidth}
        \includegraphics[width=\textwidth]{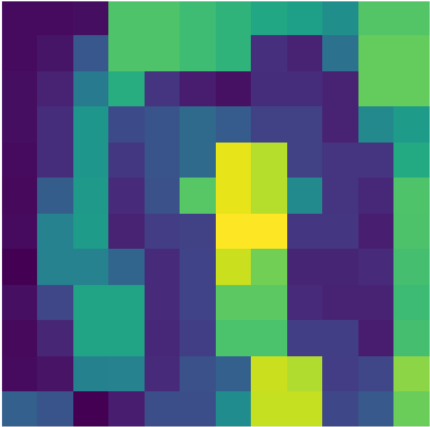}
    \end{subfigure}
    \\
    \begin{subfigure}[b]{0.22\textwidth}
        \includegraphics[width=\textwidth]{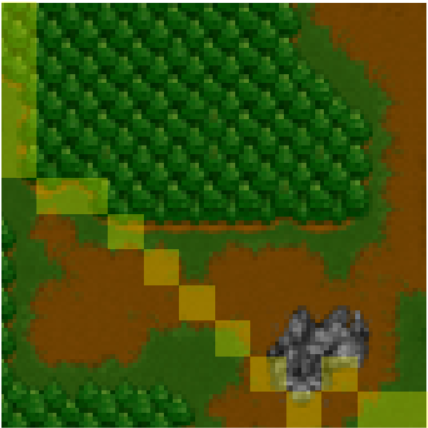}
        \caption{Initial map $x_0$ and associated shortest path $y_0$}
    \end{subfigure}
    \hfill
    \begin{subfigure}[b]{0.22\textwidth}
        \includegraphics[width=\textwidth]{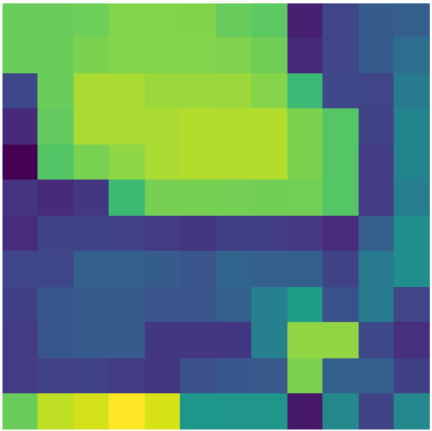}
        \caption{Initial predicted costs $\varphi(x_0)$}
    \end{subfigure}
    \hfill
    \begin{subfigure}[b]{0.22\textwidth}
        \includegraphics[width=\textwidth]{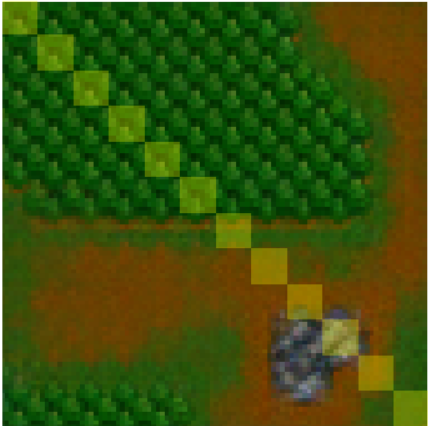}
        \caption{Counterfactual map $\xAlt$ and alternative path $\yAlt$}
    \end{subfigure}
    \hfill
    \begin{subfigure}[b]{0.22\textwidth}
        \includegraphics[width=\textwidth]{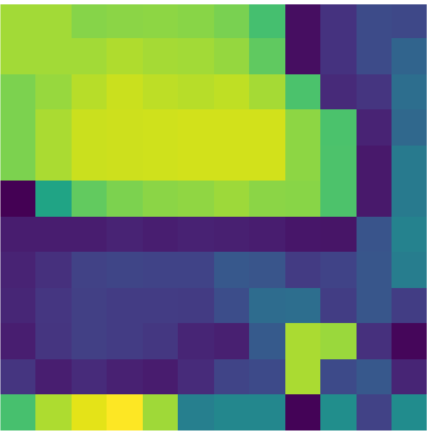}
        \caption{New predicted costs $\varphi(\xAlt)$}
    \end{subfigure}
    \caption{Example of relative explanations: explanations such that the given alternative path $\yAlt$ is shorter than the initial shortest one $y_0$ on the counterfactual map $\xAlt$.}
    \label{fig:rel_exp_examples}
\end{figure}

\begin{figure}
    \centering
    \begin{subfigure}[b]{0.22\textwidth}
        \includegraphics[width=\textwidth]{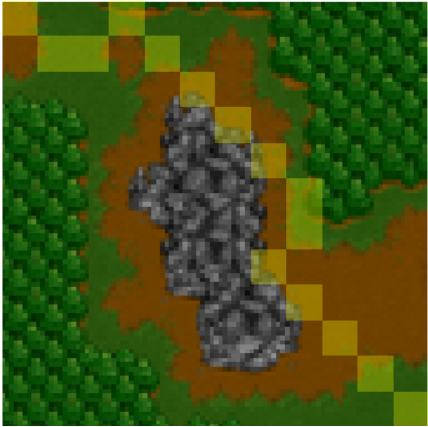}

    \end{subfigure}
    \hfill
    \begin{subfigure}[b]{0.22\textwidth}
        \includegraphics[width=\textwidth]{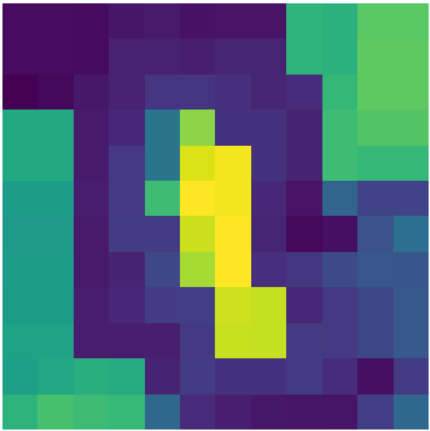}
    \end{subfigure}
    \hfill
    \begin{subfigure}[b]{0.22\textwidth}
        \includegraphics[width=\textwidth]{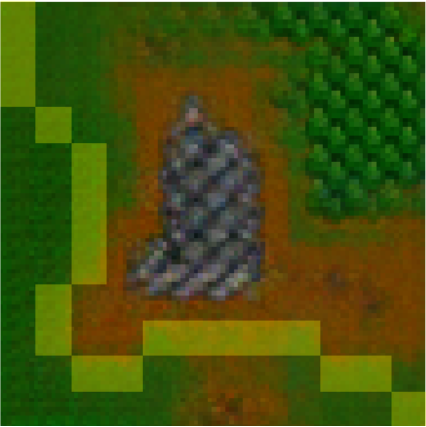}
    \end{subfigure}
    \hfill
    \begin{subfigure}[b]{0.22\textwidth}
        \includegraphics[width=\textwidth]{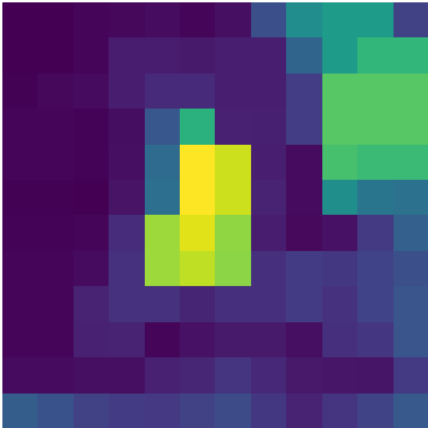}
    \end{subfigure}
    \\
    \begin{subfigure}[b]{0.22\textwidth}
        \includegraphics[width=\textwidth]{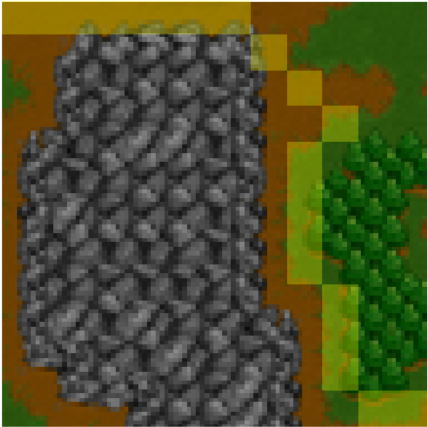}
    \end{subfigure}
    \hfill
    \begin{subfigure}[b]{0.22\textwidth}
        \includegraphics[width=\textwidth]{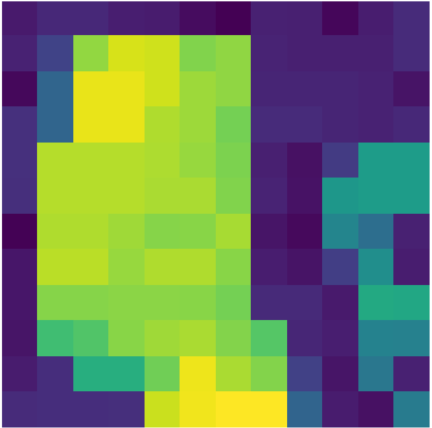}
    \end{subfigure}
    \hfill
    \begin{subfigure}[b]{0.22\textwidth}
        \includegraphics[width=\textwidth]{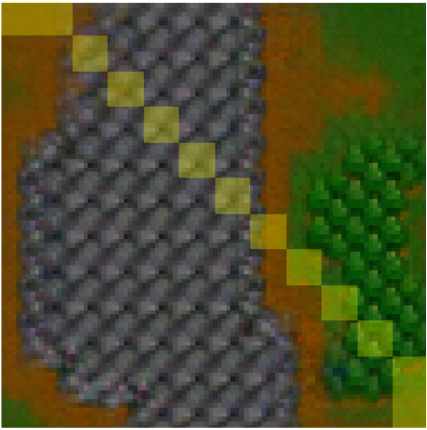}
    \end{subfigure} 
    \hfill
    \begin{subfigure}[b]{0.22\textwidth}
        \includegraphics[width=\textwidth]{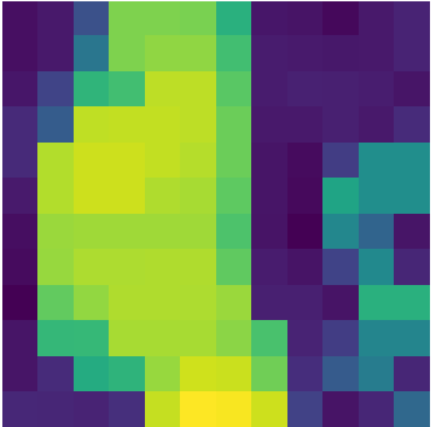}
    \end{subfigure}
    \\
    \begin{subfigure}[b]{0.22\textwidth}
        \includegraphics[width=\textwidth]{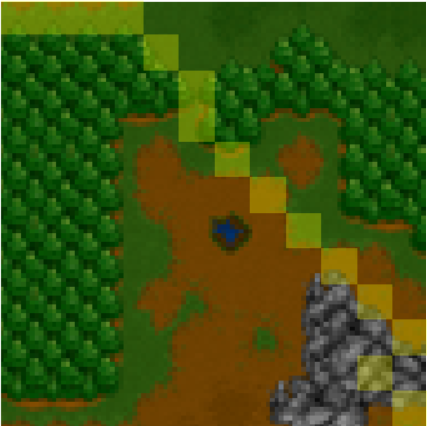}
    \end{subfigure}
    \hfill
    \begin{subfigure}[b]{0.22\textwidth}
        \includegraphics[width=\textwidth]{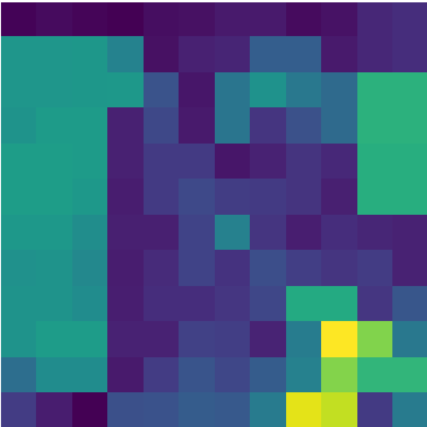}
    \end{subfigure}
    \hfill
    \begin{subfigure}[b]{0.22\textwidth}
        \includegraphics[width=\textwidth]{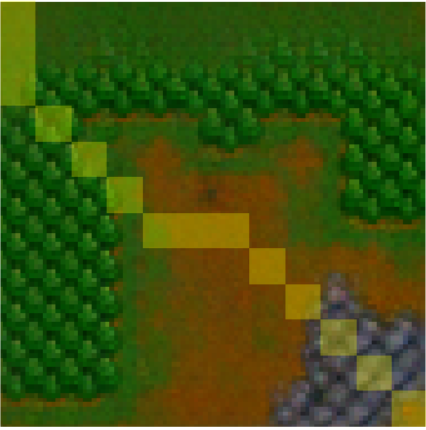}
    \end{subfigure}
    \hfill
    \begin{subfigure}[b]{0.22\textwidth}
        \includegraphics[width=\textwidth]{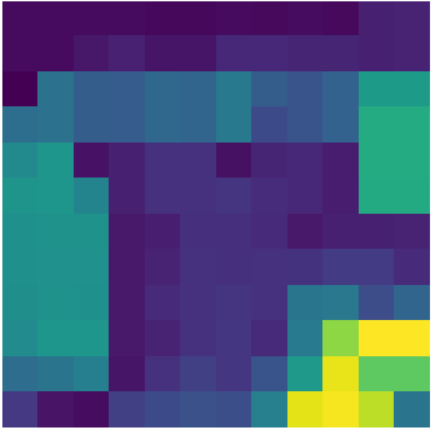}
    \end{subfigure}
    \hfill

    \begin{subfigure}[b]{0.22\textwidth}
        \includegraphics[width=\textwidth]{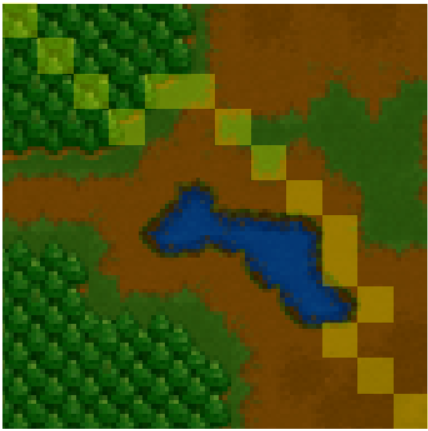}
        \caption{Initial map $x_0$ and associated shortest path $y_0$}
    \end{subfigure}
    \hfill
    \begin{subfigure}[b]{0.22\textwidth}
        \includegraphics[width=\textwidth]{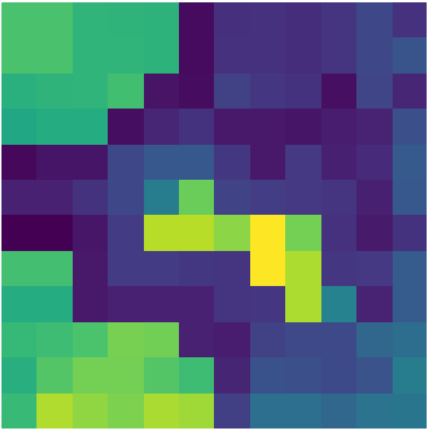}
        \caption{Initial predicted costs $\varphi(x_0)$}
    \end{subfigure}
    \hfill
    \begin{subfigure}[b]{0.22\textwidth}
        \includegraphics[width=\textwidth]{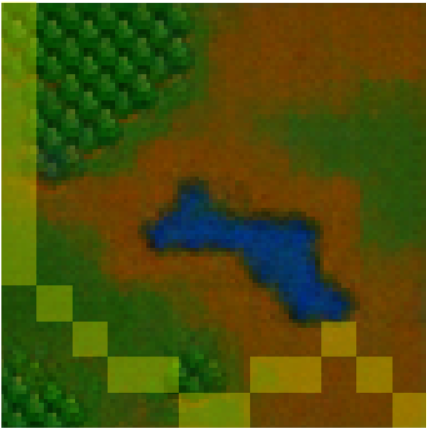}
        \caption{Counterfactual map $\xAlt$ and alternative path $\yAlt$}
    \end{subfigure}
    \hfill
    \begin{subfigure}[b]{0.22\textwidth}
        \includegraphics[width=\textwidth]{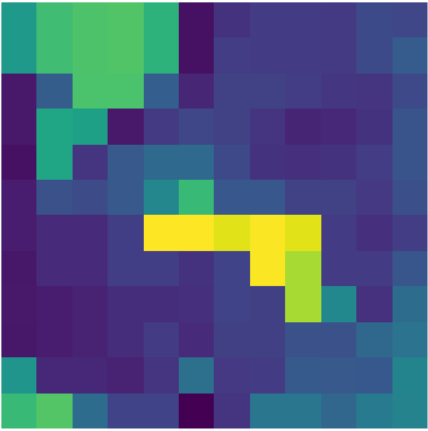}
        \caption{New predicted costs $\varphi(\xAlt)$}
    \end{subfigure}
    \caption{Example of absolute explanations: explanations such that the given alternative path $\yAlt$ is optimal on the counterfactual map~$\xAlt$.}
    \label{fig:abs_exp_examples}
\end{figure}

\begin{figure}
    \centering
    \begin{subfigure}[b]{0.22\textwidth}
        \includegraphics[width=\textwidth]{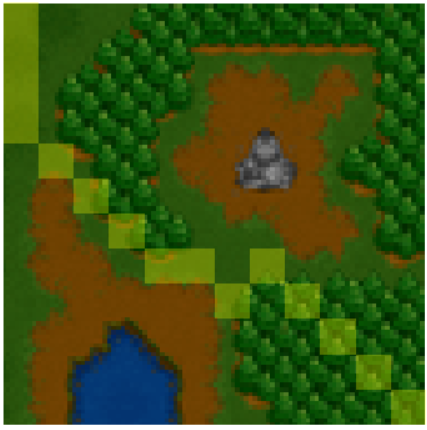}
        \caption{Initial map $x_0$ and associated shortest path $y_0$}
    \end{subfigure}
    \hfill
    \begin{subfigure}[b]{0.22\textwidth}
        \includegraphics[width=\textwidth]{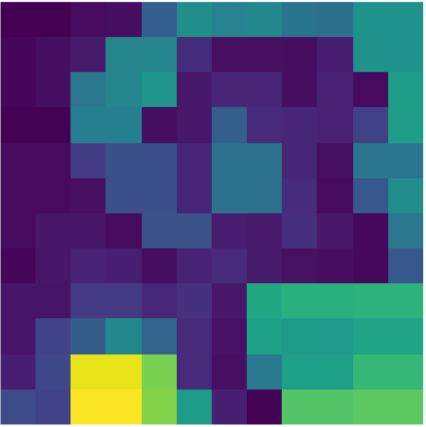}
        \caption{Initial predicted costs $\varphi(x_0)$}
    \end{subfigure}
    \hfill
    \begin{subfigure}[b]{0.22\textwidth}
        \includegraphics[width=\textwidth]{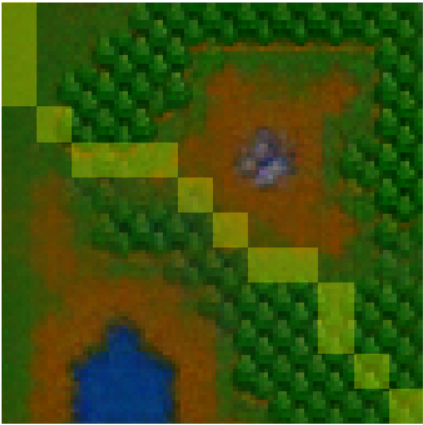}
        \caption{$\varepsilon$-explanation with $\varepsilon$=0.3 and associated shortest path}
    \end{subfigure}
    \hfill
    \begin{subfigure}[b]{0.22\textwidth}
        \includegraphics[width=\textwidth]{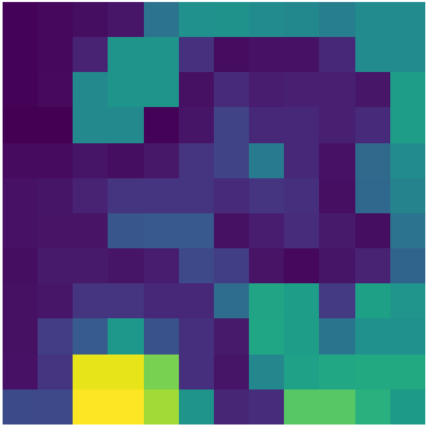}
        \caption{Corresponding predicted costs ($\varepsilon=0.3$)}
    \end{subfigure}
    \\
    \begin{subfigure}[b]{0.22\textwidth}
        \includegraphics[width=\textwidth]{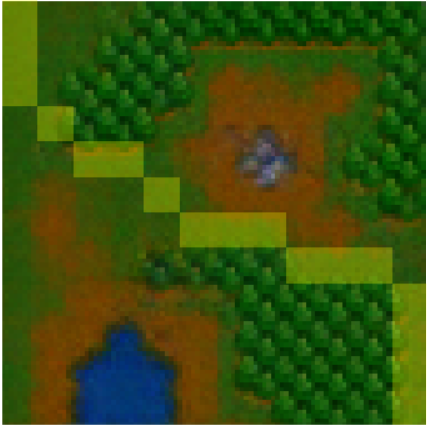}
        \caption{$\varepsilon$-explanation with $\varepsilon$=0.5 and associated shortest path}
    \end{subfigure}
    \hfill
    \begin{subfigure}[b]{0.22\textwidth}
        \includegraphics[width=\textwidth]{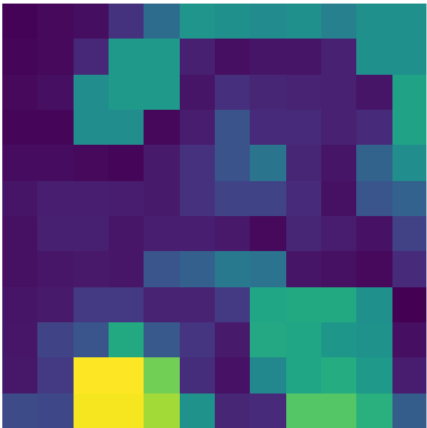}
        \caption{Corresponding predicted costs ($\varepsilon=0.5$)}
    \end{subfigure}
    \hfill
    \begin{subfigure}[b]{0.22\textwidth}
        \includegraphics[width=\textwidth]{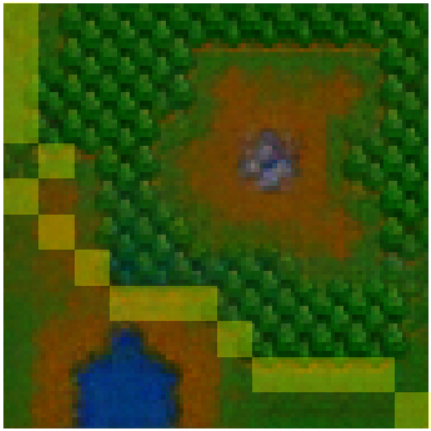}
        \caption{$\varepsilon$-explanation with $\varepsilon$=0.7 and associated shortest path}
    \end{subfigure} 
    \hfill
    \begin{subfigure}[b]{0.22\textwidth}
        \includegraphics[width=\textwidth]{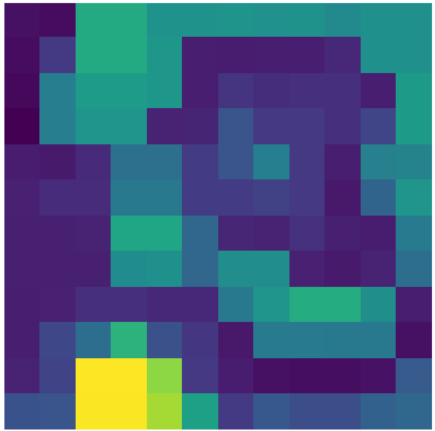}
        \caption{Corresponding predicted costs ($\varepsilon=0.7$)}
    \end{subfigure}
    \\
    \begin{subfigure}[b]{0.22\textwidth}
        \includegraphics[width=\textwidth]{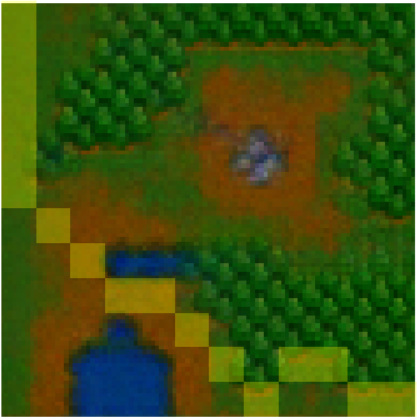}
        \caption{$\varepsilon$-explanation with $\varepsilon$=1 and associated shortest path}
    \end{subfigure}
    \hfill
    \begin{subfigure}[b]{0.22\textwidth}
        \includegraphics[width=\textwidth]{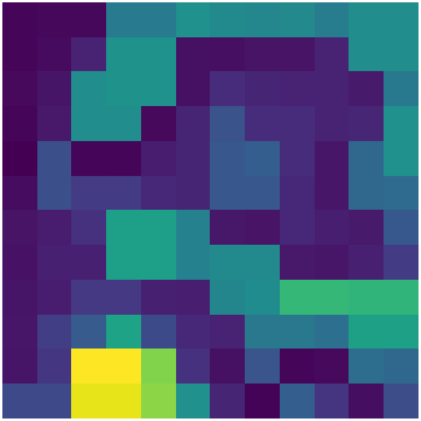}
        \caption{Corresponding predicted costs ($\varepsilon=1$)}
    \end{subfigure}
    \hfill
    \begin{subfigure}[b]{0.22\textwidth}
        \includegraphics[width=\textwidth]{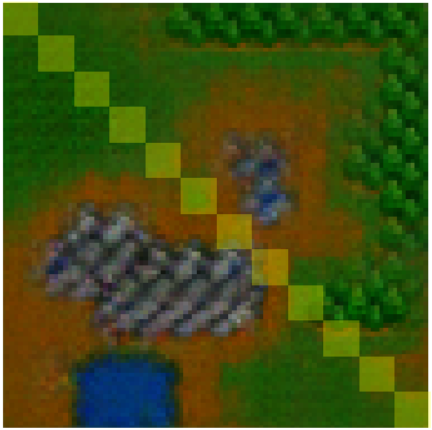}
        \caption{$\varepsilon$-explanation with $\varepsilon$=2 and associated shortest path}
    \end{subfigure}
    \hfill
    \begin{subfigure}[b]{0.22\textwidth}
        \includegraphics[width=\textwidth]{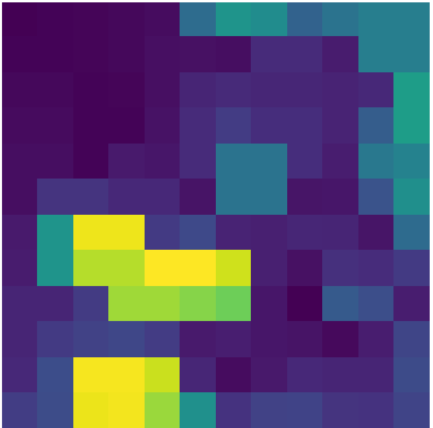}
        \caption{Corresponding predicted costs ($\varepsilon=2$)}
    \end{subfigure}
    \caption{Example of $\varepsilon$-explanations with varying $\varepsilon$: explanations such that the initial shortest path $y_0$ has relative regret of at least $\varepsilon$ on the counterfactual map $\xAlt$.}
    \label{fig:eps_exp_examples}
\end{figure}

We display several examples of counterfactual explanations for the shortest paths on Warcraft maps structured pipeline. All of them are obtained with a VAE trained in a cost-aware fashion with $\alpha=2$ and latent hypersphere regularization with $\beta=10$. We show examples of relative and absolute explanations in \cref{fig:rel_exp_examples} and \cref{fig:abs_exp_examples} respectively, and show examples of \text{$\varepsilon$-explanations} for varying $\varepsilon$ in \cref{fig:eps_exp_examples}.

Consider, for instance, the first row of \cref{fig:rel_exp_examples}. The initial map on the left shows that the shortest path goes along the gray mountain until the bottom of the map. We ask what a close Warcraft map would be, such that a path crossing the mountain diagonally (i.e., the given alternative solution $\yAlt$) is shorter than the initial one, $y_0$. The computed relative explanation, shown in the 3\textsuperscript{rd} column together with its alternative path $\yAlt$, extends the mountain to cover the bottom of the map. Notice that the rest of the image is not modified: this small change is sufficient to render the alternative path less costly than the initial one.

Similarly, the mountain is extended to the top of the image in the second row of \cref{fig:abs_exp_examples}. Again, the rest of the map is unchanged, and this small modification is enough to make the alternative diagonal path optimal (and not just shorter than the initial one, since we now ask for an absolute explanation). The explanations are not only adding mountain areas to increase travel times. They adequately capture the dataset's structure and its impact on the decision-making process. In the second row of \cref{fig:rel_exp_examples}, for instance, the explanation is obtained by removing the forest on the left edge of the map.

Counterfactual explanations provide a sensitivity analysis of the pipeline that is both local (i.e., tailored to a specific input) and decision-focused. They allow the user to contrast two decisions in the feature space and to identify what specific features drive the decision-making process.

The novel type of explanation introduced in our paper (\text{$\varepsilon$-explanations}) is especially suitable for identifying features whose small variations will most impact the decision. Consider the example given in \cref{fig:eps_exp_examples}. Figures (c) to (h) show that the central region of the map is the most sensitive: small changes in the forest blocks make the current decision sub-optimal by a factor of 30\%, 50\%, and up to 70\%. In practice, this could be used as a mechanism to tell decision-makers what features should be given special attention, for instance, to verify that their values are correctly measured.

\clearpage
\section{Supplementary Material: Algorithms}
\label{app:algo}
The latent-space and feature-space formulations of \texttt{CF-OPT} are given in \cref{alg:cf_opt_latent} and \cref{alg:cf_opt_feature} respectively.

\begin{algorithm}[htb]
    \caption{Counterfactual Explanations through First-Order Optimization - Latent Space Formulation}
    \label{alg:cf_opt_latent}
    \begin{algorithmic}
        \STATE {\bfseries Initialize:} $z^{(1)} = e_\phi\left(x_0\right)$, $\lambda^{(1)} = 0$, $c=0$, $x_{\rm{best}}=\textrm{None}$, $l_\textrm{best} = \infty$
        \FOR{$k = 1$ to $K$}
            \STATE $x^{(k)} \leftarrow d_\psi\left(z^{(k)}\right)$
            \IF{$c = c_{\max}$}
                \STATE{\bfseries Return:}   $x_{\rm{best}}$
            \ENDIF
            \STATE \texttt{/*** Check if improvement ***/}
            \IF{$x^{(k)}$ satisfies the explanation criterion}
                \IF{$\ell(x_0, x^{(k)}) + \Omega(z^{(k)}) < u \cdot l_{\rm{best}}$}
                    \STATE Store best solution:     $x_{\rm{best}} \leftarrow x^{(k)}$
                    \STATE Store best loss:     $l_{\rm{best}} \leftarrow \ell(x_0, x^{(k)}) + \Omega(z^{(k)})$
                    \STATE Reset counter: $c \leftarrow 0$
                \ELSE
                    \STATE Increase counter: $c \leftarrow c+1$
                \ENDIF
            \ENDIF
            \STATE \texttt{/*** Update variables ***/}
            \STATE Update latent code: $z^{({k+1})} \leftarrow z^{(k)} - \gamma \, \nabla_z \mathcal{E} (z^{(k)}, \lambda^{(k)})$
            \STATE Update Lagrange multiplier:
            \STATE $\lambda^{(k+1)}     \leftarrow \lambda^{(k)} + \gamma \, \nabla_\lambda   \mathcal{E}(z^{(k)}, \lambda^{(k)})$
        \ENDFOR
        \STATE{\bfseries Return:}   $x_{\rm{best}}$
    \end{algorithmic}
\end{algorithm}
\begin{algorithm}[htb]
    \caption{Counterfactual Explanations through First-Order Optimization - Feature Space Formulation}
    \label{alg:cf_opt_feature}
    \begin{algorithmic}
        \STATE {\bfseries Initialize:} $x^{(1)} = x_0$, $\lambda^{(1)} = 0$, $c=0$, $x_{\rm{best}}=\textrm{None}$
        \FOR{$k = 1$ to $K$}
            \IF{$c = c_{\max}$}
                \STATE{\bfseries Return:}   $x_{\rm{best}}$
            \ENDIF
            \STATE \texttt{/*** Check if improvement ***/}
            \IF{$x^{(k)}$ satisfies the explanation criterion}
                \IF{$\ell( x_0, x^{(k)} ) < u \cdot l_{\rm{best}}$}
                    \STATE Store best solution:     $x_{\rm{best}} \leftarrow x^{(k)}$
                    \STATE Store best loss:     $l_{\rm{best}} \leftarrow \ell( x_0,   x^{(k)})$
                    \STATE Reset counter: $c \leftarrow 0$
                \ELSE
                    \STATE Increase counter: $c \leftarrow c+1$
                \ENDIF
            \ENDIF
            \STATE \texttt{/*** Update variables ***/}
            \STATE Update covariate: $x^{({k+1})} \leftarrow x^{(k)} - \gamma \, \nabla_x \mathcal{E} (x^{(k)}, \lambda^{(k)})$
            \STATE Update Lagrange multiplier:
            \STATE $\lambda^{(k+1)}     \leftarrow \lambda^{(k)} + \gamma \, \nabla_\lambda   \mathcal{E}(x^{(k)}, \lambda^{(k)})$
        \ENDFOR
        \STATE{\bfseries Return:}   $x_{\rm{best}}$
    \end{algorithmic}
\end{algorithm}

\end{document}